\documentclass[10pt,twocolumn,letterpaper]{article}

\usepackage{iccv}
\usepackage{times}
\usepackage{epsfig}
\usepackage{graphicx}
\usepackage{amsmath}
\usepackage{amssymb}
\usepackage[accsupp]{axessibility}
\usepackage[utf8]{inputenc} 
\usepackage[T1]{fontenc}    
\usepackage{booktabs}       
\usepackage{amsfonts}       
\usepackage{nicefrac}       
\usepackage{microtype}      

\usepackage{amsmath,bbm,bm,graphicx,amsthm}
\usepackage{subfigure}
\usepackage{bbding}
\usepackage{multirow}

\DeclareMathOperator{\MSE}{MSE}
\DeclareMathOperator{\PSNR}{PSNR}
\DeclareMathOperator{\SSIM}{SSIM}
\DeclareMathOperator{\SSSIM}{S3IM}
\DeclareMathOperator{\Patch}{\mathcal{P}}
\DeclareMathOperator{\std}{std}

\usepackage[linesnumbered,ruled]{algorithm2e}
\usepackage{authblk}

\iccvfinalcopy 

\def\iccvPaperID{9880} 

\ificcvfinal\fi

\begin{document}

\title{S3IM: Stochastic Structural SIMilarity and Its Unreasonable Effectiveness\\for Neural Fields}

\author[ ]{Zeke Xie\thanks{Equal Contributions. \textit {\{xiezeke,yangxindi\}@baidu.com}.}, Xindi Yang$^{*}$, Yujie Yang,\\Qi Sun, Yixiang Jiang, Haoran Wang, Yunfeng Cai, and Mingming Sun}
\affil[ ]{Baidu Research}


\maketitle

\begin{abstract}
Recently, Neural Radiance Field (NeRF) has shown great success in rendering novel-view images of a given scene by learning an implicit representation with only posed RGB images. NeRF and relevant neural field methods (e.g., neural surface representation) typically optimize a point-wise loss and make point-wise predictions, where one data point corresponds to one pixel. Unfortunately, this line of research failed to use the collective supervision of distant pixels, although it is known that pixels in an image or scene can provide rich structural information. To the best of our knowledge, we are the first to design a nonlocal multiplex training paradigm for NeRF and relevant neural field methods via a novel Stochastic Structural SIMilarity (S3IM) loss that processes multiple data points as a whole set instead of process multiple inputs independently. Our extensive experiments demonstrate the unreasonable effectiveness of S3IM in improving NeRF and neural surface representation for nearly free. The improvements of quality metrics can be particularly significant for those relatively difficult tasks: e.g., the test MSE loss unexpectedly drops by more than $90\%$ for TensoRF and DVGO over eight novel view synthesis tasks; a $198\%$ F-score gain and a $64\%$ Chamfer $L_{1}$ distance reduction for NeuS over eight surface reconstruction tasks. Moreover, S3IM is consistently robust even with sparse inputs, corrupted images, and dynamic scenes.
\end{abstract}

\section{Introduction}
\label{sec:intro}

Synthesizing novel-view images of a 3D scene from a group of images is a long-standing task in computer vision and computer graphics \cite{chen1993view,debevec1996modeling,levoy1996light,gortler1996lumigraph,shum2000review}. This long-standing task has recently made significant progress due to advances in learning-based neural rendering methods \cite{park2019deepsdf,liu2020neural,mildenhall2021nerf}. Learning-based neural field methods can represent 3D scenes and even the corresponding surfaces from posed images toward photorealistic novel-view synthesis. 

\begin{figure}[t]
\center
\renewcommand*{\arraystretch}{0}
\begin{tabular}{c@{}c@{}c@{}c}
 & Ground Truth & Standard & Multiplex (Ours)\\
\rotatebox[origin=l]{90}{DVGO - Replica} &
\includegraphics[width =0.32\columnwidth ]{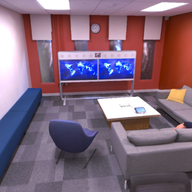} &
\includegraphics[width =0.32\columnwidth ]{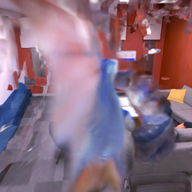}&
\includegraphics[width =0.32\columnwidth ]{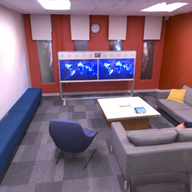}\\
\rotatebox[origin=l]{90}{DVGO - T$\&$T} &
\includegraphics[width =0.32\columnwidth ]{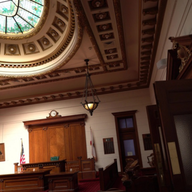} &
\includegraphics[width =0.32\columnwidth ]{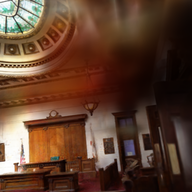}&
\includegraphics[width =0.32\columnwidth ]{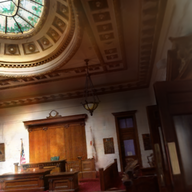}\\
\rotatebox[origin=l]{90}{NeRF - T$\&$T} &
\includegraphics[width =0.32\columnwidth ]{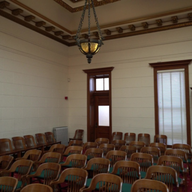} &
\includegraphics[width =0.32\columnwidth ]{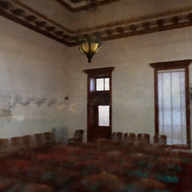}&
\includegraphics[width =0.32\columnwidth ]{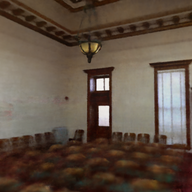}\\
\rotatebox[origin=l]{90}{TensoRF - T$\&$T} &
\includegraphics[width =0.32\columnwidth ]{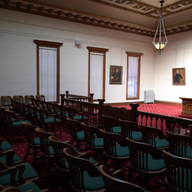} &
\includegraphics[width =0.32\columnwidth ]{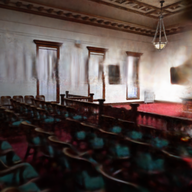}&
\includegraphics[width =0.32\columnwidth ]{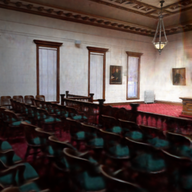}\\
\end{tabular}
\caption{Qualitative comparison of standard training and \textit{multiplex training} for neural radiance field on Replica Dataset \cite{replica19arxiv} and T$\&$T-Advanced Dataset \cite{Knapitsch2017}. Model: DVGO, TensoRF, and NeRF.}
 \label{fig:nerfqualitative}
\end{figure}

Particularly, benefited from strong representations of deep neural networks (DNNs), Neural Radiance Field (NeRF) \cite{mildenhall2021nerf} has shown impressive success in synthesizing novel view synthesis of a given scene by implicitly encoding volumetric density and color through a fully connected neural network (often referred to as a multilayer perceptron or MLP). NeRF regresses from a single 5D representation $(x, y, z, \theta, \phi)$- 3D coordinates $\bm{x}=(x, y, z)$ plus 2D viewing directions $\bm{d}=(\theta, \phi)$- a single volume density $\sigma$ and a view-dependent RGB color $\bm{c}=(r,g,b)$, computed by fitting the model to a group of pixels (from training images). NeRF approximates this continuous 5D scene representation with an MLP network $f_{\Theta}: (\bm{x}; \bm{d}) \rightarrow (\bm{c};\sigma)$ and optimizes its weights $\Theta$ to map each input 5D coordinate to the corresponding volume density and directional emitted color.

Without loss of generality, we focus on the learning module of NeRF and write the loss optimized by NeRF as
\begin{align}
\label{eq:simplexloss}
L (\Theta) = \frac{1}{\|\mathcal{R}\|}\sum_{\bm{r}\in \mathcal{R}} l_{\MSE}(\Theta, \bm{r}),
\end{align}
where $l_{\MSE}$ is the point-wise Mean Squared Error (MSE) loss for each pixel/ray $\bm{r} = (\bm{x}_{\bm{r}}, \bm{d}_{\bm{r}}, \bm{c}_{\bm{r}})$ in the training data or data minibatch $\mathcal{R}$. Obviously, the MLP of NeRF learns and makes inference point-wisely, because one data point of the MLP corresponds to one pixel's information. 

NeRF and neural surface representation are two of the most representative neural field methods \cite{xie2022neural}, which are also called implicit neural representation methods \cite{sitzmann2020implicit} in computer vision. Neural surface representation \cite{michalkiewicz2019implicit,niemeyer2020differentiable,yariv2021volume,wang2021neus} is another important and long-standing problem orthogonal to NeRF. The point-wise MSE loss not only measures the accuracy of predictions in NeRF, but is also widely used in other relevant neural field methods. However, only optimizing point-wise loss can be a serious but overlooked pitfall of training NeRF and relevant methods.

In image quality assessment, the point-wise MSE-based metric Peak Signal-to-Noise Ratio (PSNR) \cite{eskicioglu1995image} is widely used but do not correlate well with perceived image quality \cite{pappas2000perceptual,wang2002universal}, since point-wise metrics do not reflect structural information \cite{koenderink1984structure,wang2004image,wang2003multiscale,sheikh2006image} that is rich in the physical world.

The Structural Similarity (SSIM) index not only reflects the structure of a group of pixels, but also correlates with human visual systems significantly better than PSNR/MSE \cite{wang2004image,hore2010image}. SSIM is an important performance metric for evaluating NeRF models, but it is not used for training NeRF models. The conventional training paradigm of NeRF has unfortunately overlooked the structural information, as it only optimizes the point-wise MSE loss but not using the collective supervision of the structural information of multiple pixels. This is not surprising. The conventional point-wise paradigm is also common in other machine learning models.

Can we propose a novel training paradigm that can capture the structural information of a group of inputs in ways other than the MSE loss of individual pixels? Yes, we formulate a \textit{multiplex loss} associated with the novel training paradigm as
\begin{align}
\label{eq:multiplexloss}
L_{\mathrm{M}} (\Theta) = \frac{1}{\|\mathcal{R}\|}\sum_{\bm{r}\in \mathcal{R}} l_{\MSE}(\Theta, \bm{r}) + \lambda L_{\SSSIM}(\Theta, \mathcal{R}),
\end{align}
where $L_{\SSSIM}(\Theta, \mathcal{R})$ is computed over a group of nonlocal pixels and the hyperparameter $\lambda$ adjusts the importance of $\SSSIM$. Unlike $L_{\SSSIM}(\theta, \mathcal{R})$, the conventional loss $l_{\MSE}(\theta, (\bm{x}, \bm{d}, \bm{c}, \sigma))$ is computed pixel-wisely. We emphasize that $L_{\SSSIM}(\Theta, \mathcal{R})$ cannot be expressed as the sum of any other loss computed over individual pixels from $\mathcal{R}$. We call the proposed $L_{\SSSIM}$ a multiplex loss because it allows a model to process multiple nonlocal inputs as a whole multiplex input. We will formally define S3IM in Section \ref{sec:method}.

\textbf{Main Contributions.} We summarize main contributions as follows. \textbf{(1)} We propose the novel Stochastic Structural SIMilarity (S3IM) index, which measures the similarity between two groups of pixels and captures nonlocal structural similarity information from stochastically sampled pixels. 
To the best of our knowledge, we are the first to formulate a multiplex loss that can process multiple inputs collectively by capturing nonlocal information rather than processing multiple inputs independently by capturing individual-pixel information in neural fields. \textbf{(2)} S3IM is model-agnostic and can be generally applied to all types of neural field methods, such as NeRF and neural surface representation, with limited coding costs and computational costs. Our extensive experiments demonstrate the unreasonable effectiveness of the \textit{nonlocal multiplex training} paradigm in improving NeRF and neural surface representation. Particularly, the improvement in quality metrics can be especially significant for those difficult tasks (e.g., Tables \ref{table:dvgoreplica} and \ref{table:neusreplica} and Figure \ref{fig:nerfqualitative}) and is robust even with sparse inputs, corrupted images, and dynamic scenes.

\section{Background}
\label{sec:backgound}

In this section, we introduce background knowledge.

\subsection{NeRF}

Recall that NeRF maps from a single 5D representation $(x, y, z, \theta, \phi)$ to a single volume density $\sigma$ and view-dependent RGB color $\bm{c}=(r,g,b)$ with an MLP network: $f_{\Theta}: (\bm{x}; \bm{d}) \rightarrow (\bm{c};\sigma)$. For a target view with pose, a camera ray can be parameterized as $\bm{r}(t) = \bm{o} + t\bm{d}$ with the ray origin $\bm{o}$ and ray unit direction $\bm{d}$. The expected color $\bm{C}(\bm{r})$ of camera ray $\bm{r}(t)$ with near and far bounds $t_n$ and $t_f$ is
\begin{align}
\label{eq:volrender}
 \hat{\bm{C}}(\bm{r}) = \int_{t_{n}}^{t_{f}} T(t) \sigma(t) \bm{c}(t) dt,
\end{align}
where $T = \exp(-\int_{t_n}^{t} \sigma(s) ds)$ denotes the accumulated transmittance along the ray from $t_n$ to $t$. For simplicity, we have ignored the coarse and fine renderings via different sampling methods.

The rendered image pixel value for camera ray $\bm{r}$ can then be compared against the corresponding ground truth pixel
value $\bm{C}(\bm{r})$, for all the camera rays. The conventional NeRF rendering loss is the MSE loss
\begin{align}
\label{eq:mseloss}
L (\Theta) = \frac{1}{\|\mathcal{R}\|}\sum_{\bm{r}\in \mathcal{R}} \|  \hat{\bm{C}}(\bm{r}) - \bm{C}(\bm{r}) \|^{2}.
\end{align}
Obviously, NeRF is a point-wise machine learning model during both training and inference. The prediction and the loss are both computed pixel-wisely.

\subsection{Quality Metrics: PSNR and SSIM}

PSNR and SSIM are two most popular metrics of image quality assessment \cite{wang2004image,hore2010image}. For simplicity, we take grey-level (8 bits) images as examples. Given a test image $I_{a}$ and a reference image $I_{b}$, both of size $W\times H $, the PSNR can be defined as
\begin{align}
\label{eq:psnr}
\PSNR(I_{a}, I_{b}) = 10 \log_{10} \left( \frac{255^2}{\MSE(I_{a}, I_{b})} \right),
\end{align}
where $\MSE(I_{a}, I_{b}) = \frac{1}{WH} \sum_{i,j,k} (I_{b,ijk}- I_{a,ijk})^{2}. $
It is easy to see that PSNR directly depends on MSE and overlooks the collective information of a group of pixels. 

In contrast, SSIM is a well-known quality metric that can capture local structural similarity between images or patches. SSIM is considered to be correlated with the quality perception of the human visual system well and is widely used for evaluating NeRF \cite{wang2004image,hore2010image}. Suppose $\bm{a} = \{a_{i} | i=1,2,3, \dots, n\}$ and $\bm{b} = \{b_{i} | i=1,2,3, \dots, n\}$ to be
two discrete non-negative signals paired with each other (e.g., two image patches extracted from the same spatial location from paired images). SSIM is expressed by the combination of three terms which are the luminance, contrast, and structure comparison metrics:
\begin{align}
\label{eq:ssim}
\SSIM(a, b) = l(\bm{a}, \bm{b}) c(\bm{a}, \bm{b}) s(\bm{a}, \bm{b}).
\end{align}
The luminance $l(\bm{a}, \bm{b})$, contrast $c(\bm{a}, \bm{b})$, and structure comparison $s(\bm{a}, \bm{b})$ are, respectively, written as 
\begin{align}
\label{eq:lcs}
l(\bm{a}, \bm{b}) =& \frac{2\mu_{a}\mu_{b} + C_{1}}{\mu_{a}^{2} + \mu_{b}^{2} + C_{1}},\\
c(\bm{a}, \bm{b}) =& \frac{2\sigma_{a}\sigma_{b} + C_{2}}{\sigma_{a}^{2} + \sigma_{b}^{2} + C_{2}},\\
s(\bm{a}, \bm{b}) =& \frac{\sigma_{ab} + C_{3}}{\sigma_{a}\sigma_{b}  + C_{3}}.
\end{align}
where $C_1$, $C_2$, and $C_3$ are small constants given by 
$C_1 = (K_{1}L)^{2}\text{, }C_2 = (K_{2}L)^{2}\text{, and }C_3= C_2 / 2.$
Following the common setting \cite{wang2004image,mildenhall2021nerf}, we have $K_{1}=0.01$, $K_{2}=0.03$, and $L=1$ for RGB. The range of SSIM lies in $[-1,1]$. 

In practice of image quality assessment, people usually apply the SSIM index locally rather than globally for robust statistics and efficient computation. The local statistics, including mean $\mu_{a}$, variance $\sigma_{a}$, and covariance $\sigma_{ab}$ are computed within a local $K\times K$ kernel window, which moves with a stride size $s$ over the entire image. At each step, the local statistics and SSIM index are calculated within the local window. The final SSIM metric for evaluating NeRF is actually the mean SSIM (MSSIM) which is computed by averaging the SSIM indexes over each step. We leave the more details of SSIM in Appendix B.

\section{Methodology: Multiplex Training via S3IM}
\label{sec:method}

In this section, we formulate a novel multiplex loss, S3IM, with a novel model-agnostic multiplex training paradigm for neural fields.

\subsection{S3IM}
\label{sec:methodnerf}

Our motivation is to let the collective and nonlocal structural information contained in the group of data points supervise the learning of neural fields. Therefore, we must first define a multiplex-style loss over a group of pixels that can capture structural information, rather than the conventional point-wise loss (e.g., MSE) defined over individual pixels.

The pixels in a local patch may contain certain positional information. However, due to the stochastic training, the stochastically sampled pixels in a minibatch $\mathcal{R}$ cannot form a local patch and lose the positional relationship completely. 

We formulate a stochastic variant of SSIM, namely S3IM, for stochastic training of NeRF. The idea is concise. Suppose we have $B$ (e.g., 1024) pixels per minibatch and choose the kernel size and the stride size of S3IM as $K\times K$ (e.g., $4\times 4$) and $s=K$. For simplicity and efficiency, we choose the stride size to be the same as the kernel size because the stochastic patches from one minibatch are independent and without overlapping pixels in this case.

\begin{algorithm}[t]
\label{algo:multiplex}
\caption{Multiplex Training via S3IM}
\SetKwInput{KwInput}{Input}               
\SetKwInput{KwOutput}{Output}  
  Let $\mathcal{A}$ be an SGD-like training algorithm\; 
  \While{no stopping criterion has been met}
  { 
   Sample a data minibatch of rays $\mathcal{R}$ from $\mathcal{D}$\;
   Obtain the ground-truth pixels $\mathcal{C} = \{\bm{C}(\bm{r})|\bm{r}\in \mathcal{R}\}$\;
   Compute the rendered pixels $\hat{\mathcal{C}} = \{\hat{\bm{C}}(\bm{r})|\bm{r}\in \mathcal{R}\}$\; 
     \For{$m=1$ \KwTo $M$}
     {  
    Initialize the stochastic patch generation function $\Patch^{(m)}$\;
    Transform the rendered pixels into the rendered stochastic patch $\Patch^{(m)}(\hat{\mathcal{C}})$\; 
    Transform the ground-truth pixels into the ground-truth stochastic patch $\Patch^{(m)}(\mathcal{C})$\;
    Compute $\SSIM(\Patch^{(m)}(\hat{\mathcal{C}}), \Patch^{(m)}(\mathcal{C}))$ with the given kernel $K \times K$ and the stride size $K$\;
     }
    Obtain the S3IM loss $L_{\SSSIM}(\Theta) =1- \frac{1}{M} \sum_{m=1}^{M} \SSIM(\Patch^{(m)}(\hat{\bm{C}}(\mathcal{R})), \Patch^{(m)}(\bm{C}(\mathcal{R})))$\;
    Obtain the conventional MSE loss $L (\Theta) = \frac{1}{\|\mathcal{R}\|}\sum_{\bm{r}\in \mathcal{R}} \|  \hat{\bm{C}}(\bm{r}) - \bm{C}(\bm{r}) \|^{2}$\;
    $L_{M}(\Theta) = L (\Theta) + \lambda L_{\SSSIM}(\Theta)  $\;
    Compute the gradient $\nabla L_{\mathrm{M}} (\Theta)$\;
    Update the model parameters $\Theta$ by $\mathcal{A}$\;
  }
\end{algorithm}

We summarize S3IM as three steps: 
\vspace{-0.5em}
\begin{enumerate}
\item We let $B$ rays/pixels from a dataset/minibatch $\mathcal{R}$ randomly form a rendered patch $\Patch(\hat{\mathcal{C}})$ and the corresponding ground-truth image patch $\Patch(\mathcal{C})$, where $\hat{\mathcal{C}} = \{\hat{\bm{C}}(\bm{r})|\bm{r}\in \mathcal{R}\}$ and $\mathcal{C} = \{\bm{C}(\bm{r})|\bm{r}\in \mathcal{R}\}$. 
\vspace{-0.5em}
\item We compute SSIM with the kernel size $K\times K$ and the stride size $s$ over the rendered patch and the corresponding ground-truth patch, which is exactly an estimator of the proposed S3IM over the paired stochastic patches.
\vspace{-0.5em}
\item Due to stochasticity of $\Patch(\cdot)$, we may repeat steps \textbf{(1)} and \textbf{(2)} $M$ times and average the $M$ estimated SSIM values to obtain the final S3IM.
\end{enumerate}
\vspace{-0.5em}
In summary, the final S3IM can be written as
\begin{align}
\label{eq:s3im}
 \SSSIM(\hat{\mathcal{R}},\mathcal{R})
= \frac{1}{M} \sum_{m=1}^{M} \SSIM(\Patch^{(m)}(\hat{\mathcal{C}}), \Patch^{(m)}(\mathcal{C})),
\end{align}
where SSIM needs to apply the kernel size $K\times K$ and the stride size $s$.

We note that computing S3IM can be well vectorized and \textit{multiplex training} also only requires back-propagation for once per iteration. Thus, the extra computational cost of \textit{multiplex training} is limited. As S3IM lies in $[-1,1]$ and positively correlated with image quality, we define the S3IM-based loss $L_{\SSSIM}$ as
\begin{align}
\label{eq:s3imloss}
L_{\SSSIM}(\Theta, \mathcal{R}) =& 1 - \SSSIM(\hat{\mathcal{R}},\mathcal{R}), \\
=& 1-\frac{1}{M} \sum_{m=1}^{M} \SSIM(\Patch^{(m)}(\hat{\mathcal{C}}), \Patch^{(m)}(\mathcal{C})). \nonumber
\end{align}

We present the pseudocode in Algorithm \ref{algo:multiplex}; for generality, we focus on the shared machine learning module of various neural field methods and ignore the details of sampling in Equation \eqref{eq:volrender}. Fortunately, similar to SSIM, we may directly apply a default setting that $K=4$ and $S=K$ without fine-tuning. The multiplex training paradigm via S3IM brings in two extra hyperparameters $\lambda$ and $M$. If we let $M$ approach $+\infty$, we will eliminate the stochasticity of S3IM and obtained its expected value. According to our experimental results in Section \ref{sec:empirical}, we observe that $M=1$ usually produce nearly same good results as $M=10$, while we choose the default value $M=10$ mainly for reducing the stochasticity of the final results. Thus, only the S3IM loss weight $\lambda$ requires fine-tuning in practice.

 While the original SSIM is also differentiable, directly optimizing it may not work well. Even if we use a pixel data loader that provides a minibatch of pixels per iteration, and an additional local-patch dataloader, which yields a minibatch of local patches per iteration, our ablation study shows that optimizing S3IM (via stochastic patches) significantly outperforms optimizing SSIM (via local patches). This is not surprising. SSIM over local patches can only capture structural information carried by nearby pixels from one image, while S3IM over stochastic patches can capture nonlocal structural information carried by distant pixels from different images. The additional data loader not only brings in extra coding and computational costs, but also hurts the performance of NeRF.

\subsection{S3IM for Neural Surface Representations}
\label{sec:methodsurface}

In this subsection, we show that it is very easy to apply the proposed \textit{multiplex training} to other neural field methods. 

NeuS \cite{wang2021neus} is a recent powerful neural surface representation method. We choose neural surface representation as another benchmark task in this paper because neural surface representation is another line of research which has received much attention and made great progress recently in computer vision and computer graphics.

NeuS optimize a color loss with a regularizer loss and an optional mask loss, which can be written as
\begin{align}
\label{eq:neusloss}
L_{\mathrm{NeuS}} = L_{\mathrm{color}} + \lambda_{\mathrm{reg}} L_{\mathrm{reg}}, 
\end{align}
where the color loss use the $L_1$ loss form as
\begin{align}
 L_{\mathrm{color}} = \frac{1}{\|\mathcal{R}\|}\sum_{\bm{r}\in \mathcal{R}} \|  \hat{\bm{C}}(\bm{r}) - \bm{C}(\bm{r}) \|_{1}.
\end{align}
Similarly, when we train NeuS in the proposed \textit{multiplex training} paradigm, we replace the original color loss by the multiplex color loss via S3IM as
\begin{align}
 L_{\mathrm{M},\mathrm{color}} =  L_{\mathrm{color}}  + \lambda L_{\SSSIM}.
\end{align}
Our neural surface reconstruction experiments demonstrate that S3IM not only improves RGB image quality metrics, but also improves other geometric quality metrics (e.g., Chamfer Distance ) even more significantly. This suggests that S3IM is generally useful for neural fields.

\section{Related Work}
\label{sec:related}

In this section, we review representative related works and discuss their relations to our method.

\textbf{Neural Fields.} Fields can continuously parameterize an underlying physical quantity of an
object or scene over space and time. Since a long time ago, fields have been used to describe physical phenomena \cite{sabella1988rendering}, compute image gradients \cite{schluns1997local}, simulate collisions \cite{osher2004level}. Recent advances showed increased interest in employing coordinate-based neural networks to parameterize some physical quantities over space and time, such as a neural network that maps a 3D spatial coordinate to a flow field in fluid dynamics, or a colour and density field in 3D scene representation. Such networks are often referred to as neural fields \cite{xie2022neural}. The application of neural fields in visual computing has make remarkable progress on various computer vision problems such as 3D scene reconstruction and generative modelling. In computer vision, people often refer to neural fields as implicit neural representations  \cite{sitzmann2020implicit,michalkiewicz2019implicit,niemeyer2020differentiable,yariv2021volume,wang2021neus} (e.g., neural surface representation).

\textbf{Neural Radiance Fields and Neural Surface Representations.} Neural radiance field \cite{mildenhall2021nerf} and neural surface representation are two of the most representative neural field methods. This is why we let NeRF and NeuS serve as the base models for evaluating the proposed S3IM and the resulted \textit{multiplex training} paradigm. 
 
 The line of NeRF has developed a number of useful NeRF variants. NeRF++ \cite{zhang2020nerf++} helped resolve the shape-radiance ambiguity of NeRF. Mip-NeRF \cite{barron2021mip} adopted a multiscale representation method and significantly improved the quality of representating fine details. 
 D-NeRF \cite{pumarola2021d} extends neural radiance fields to modeling dynamical scences. 
 Some works, such as Pixel-NeRF \cite{yu2021pixelnerf} and Reg-NeRF \cite{niemeyer2022regnerf}, focused on view synthesis from sparse inputs. NeRF$--$ \cite{wang2021nerf} performs view synthesis by estimating approximate camera poses rather than known camera poses. 
 As the common NeRF methods suffers a lot from slow training and inference, some works, including DVGO \cite{sun2022direct}, TensoRF \cite{chen2022tensorf}, and Instant NGP\cite{muller2022instant}, aimed at accelerating training and inference of NeRF. Inspired by NeRF, NeuS \cite{wang2021neus} related the occupancy function of a volume to its volume density, thereby leading to improved rendering results and better geometry reconstruction. The proposed \textit{multiplex training} paradigm via S3IM, orthogonal to the existing point-wise training paradigm, is model-agnostic and orthogonal to these NeRF variants. 
 


\section{Empirical Analysis and Discussion}
\label{sec:empirical}

\begin{table}[h]
\caption{Quantitative results of NeRF methods on Replica Dataset \cite{replica19arxiv}. Model: DVGO.}
\label{table:dvgoreplica}
\begin{center}
\begin{small}
\resizebox{0.45\textwidth}{!}{%
\begin{tabular}{c|c|ccc}
\toprule
Scene & Training &  PSNR($\uparrow$) & SSIM($\uparrow$) & LPIPS($\downarrow$)\\
\midrule 
\multirow{2}{*}{Scene 1} & Standard & 13.26 & 0.506 & 0.719 \\
  & Multiplex & \textbf{32.63} & \textbf{0.929} & \textbf{0.0685}  \\   
\midrule 
\multirow{2}{*}{Scene 2} & Standard & 14.82 & 0.653 & 0.637 \\
  & Multiplex & \textbf{35.03} & \textbf{0.957} & \textbf{0.0527}  \\   
\midrule 
\multirow{2}{*}{Scene 3} & Standard & 15.24 & 0.644 & 0.636 \\
  & Multiplex & \textbf{29.88} & \textbf{0.957} & \textbf{0.0639}  \\   
\midrule 
\multirow{2}{*}{Scene 4} & Standard & 17.73 & 0.691 & 0.505 \\
  & Multiplex & \textbf{39.32} & \textbf{0.976} & \textbf{0.0325}  \\   
\midrule 
\multirow{2}{*}{Scene 5} & Standard & 16.52 & 0.659 & 0.505 \\
  & Multiplex & \textbf{35.70} & \textbf{0.969} & \textbf{0.0589} \\   
\midrule 
\multirow{2}{*}{Scene 6} & Standard & 20.10 & 0.843 & 0.309 \\
  & Multiplex & \textbf{29.27} & \textbf{0.947} & \textbf{0.0944} \\  
\midrule 
\multirow{2}{*}{Scene 7} & Standard & 23.20 & 0.845 & 0.248 \\
  & Multiplex & \textbf{31.53} & \textbf{0.952} & \textbf{0.0648} \\ 
\midrule 
\multirow{2}{*}{Scene 8} & Standard & 15.67 & 0.729 & 0.521 \\
  & Multiplex & \textbf{34.66} & \textbf{0.956} & \textbf{0.0740} \\  
\midrule 
\multirow{2}{*}{Mean} & Standard & 17.07 & 0.696 & 0.510 \\
  & Multiplex & \textbf{33.50} & \textbf{0.955} & \textbf{0.0637} \\ 
\bottomrule
\end{tabular}
}
\end{small}
\end{center}
\end{table}

\begin{table}[h]
\caption{Quantitative results of NeRF methods on Replica Dataset \cite{replica19arxiv}. Model: TensoRF.}
\label{table:tensorfreplica}
\begin{center}
\begin{small}
\resizebox{0.45\textwidth}{!}{%
\begin{tabular}{c|c|ccc}
\toprule
Scene & Training &  PSNR($\uparrow$) & SSIM($\uparrow$) & LPIPS($\downarrow$)\\
\midrule 
\multirow{2}{*}{Scene 1} & Standard & 12.13 & 0.468 & 0.7786 \\
  & Multiplex & \textbf{37.15} & \textbf{0.958} & \textbf{0.0335} \\ 
\midrule 
\multirow{2}{*}{Scene 2} & Standard & 14.17 & 0.605 & 0.709 \\
  & Multiplex & \textbf{36.55} & \textbf{0.952} & \textbf{0.0563} \\ 
\midrule 
\multirow{2}{*}{Scene 3} & Standard & 15.20 & 0.642 & 0.695 \\
  & Multiplex & \textbf{38.79} & \textbf{0.977} & \textbf{0.0287} \\ 
\midrule 
\multirow{2}{*}{Scene 4} & Standard & 19.63 & 0.625 & 0.659 \\
  & Multiplex & \textbf{44.65} & \textbf{0.990} & \textbf{0.00965} \\ 
\midrule 
\multirow{2}{*}{Scene 5} & Standard & 19.63 & 0.638 & 0.525 \\
  & Multiplex & \textbf{40.96} & \textbf{0.980} & \textbf{0.0273} \\ 
\midrule 
\multirow{2}{*}{Scene 6} & Standard & 10.92 & 0.507 & 0.693 \\
  & Multiplex & \textbf{37.94} & \textbf{0.969} & \textbf{0.0473} \\ 
\midrule 
\multirow{2}{*}{Scene 7} & Standard & 11.06 & 0.475 & 0.751 \\
  & Multiplex & \textbf{36.77} & \textbf{0.966} & \textbf{0.0437} \\ 
\midrule 
\multirow{2}{*}{Scene 8} & Standard & 11.73 & 0.630 & 0.797 \\
  & Multiplex & \textbf{39.59} & \textbf{0.977} & \textbf{0.0307} \\ 
\midrule 
\multirow{2}{*}{Mean} & Standard & 14.30 & 0.574 & 0.689 \\
  & Multiplex & \textbf{39.05} & \textbf{0.971} & \textbf{0.0454} \\ 
\bottomrule
\end{tabular}
}
\end{small}
\end{center}
\end{table}

In this section, we empirically demonstrate that \textit{multiplex training} via S3IM significantly outperform the conventional training paradigm for NeRF and its variants. 

We let the experimental settings follow original papers to produce the baselines, unless we specify otherwise. The main principle of our experimental setting is to fairly compare \textit{multiplex training} via S3IM and standard training for NeRF and relevant neural field methods. Thus, we keep all hyperparameters same for standard training and \textit{multiplex training} except S3IM. We mainly used five recent representative methods, including, vanilla NeRF \cite{mildenhall2021nerf}, DVGO \cite{sun2022direct}, TensoRF \cite{chen2022tensorf}, D-NeRF \cite{pumarola2021d}, and NeuS \cite{wang2021neus}. We present the experimental details in Appendix A.

\subsection{Novel View Synthesis Experiments}

\begin{table}[h]
\caption{Quantitative results of NeRF methods on T$\&$T. The mean PSNR, SSIM, LPIPS are computed over four scenes of T$\&$T. Model: DVGO, TensoRF, and (vanilla) NeRF.}
\vspace{-0.2in}
\label{table:nerfmeantntad}
\begin{center}
\begin{small}
\resizebox{0.48\textwidth}{!}{%
\begin{tabular}{c|c|ccc}
\toprule
Model & Training &  PSNR($\uparrow$) & SSIM($\uparrow$) & LPIPS($\downarrow$)\\
 \midrule 
\multirow{2}{*}{DVGO}  & Standard & 22.42 & 0.776 & 0.236 \\
 & Multiplex & \textbf{23.20} & \textbf{0.809} & \textbf{0.176} \\  
 \midrule 
\multirow{2}{*}{TensoRF}  & Standard & 19.69 & 0.650 & 0.365 \\
 & Multiplex & \textbf{22.85} & \textbf{0.777} & \textbf{0.230} \\  
 \midrule 
\multirow{2}{*}{NeRF}  & Standard & 21.02 & 0.659 & 0.364 \\
 & Multiplex & \textbf{22.64} & \textbf{0.725} & \textbf{0.304} \\
\bottomrule
\end{tabular}
}
\end{small}
\end{center}
\end{table}

\textbf{Static scene synthesis.} We first study how \textit{multiplex training} via S3IM improves NeRF on two benchmark datasets, Replica Dataset \cite{replica19arxiv} and Tanks and Temples Advanced (T$\&$T) \cite{Knapitsch2017}. Replica Dataset is a relatively difficult dataset which include large-scale scenes with less training images. T$\&$T is a popular benchmark dataset for image-based 3D reconstruction with more training images. We use T$\&$T Advanced as the defaulted T$\&$T in the main text as it contain more complex details, while we leave the experimental results of T$\&$T Intermediate in Appendix C.

\begin{figure*}[h]
\center
\renewcommand*{\arraystretch}{0}
\begin{tabular}{c@{}c@{}c@{}c}
 & Ground Truth & Standard & Multiplex (Ours)\\
\rotatebox[origin=l]{90}{\parbox{0.9in}{\centering Sparse}} &
\includegraphics[width =0.66\columnwidth ]{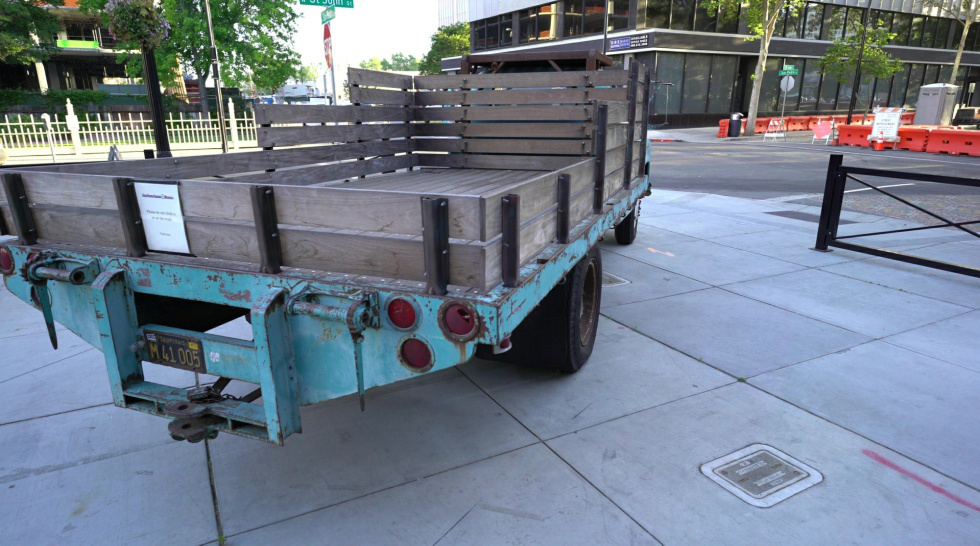}&
\includegraphics[width =0.66\columnwidth ]{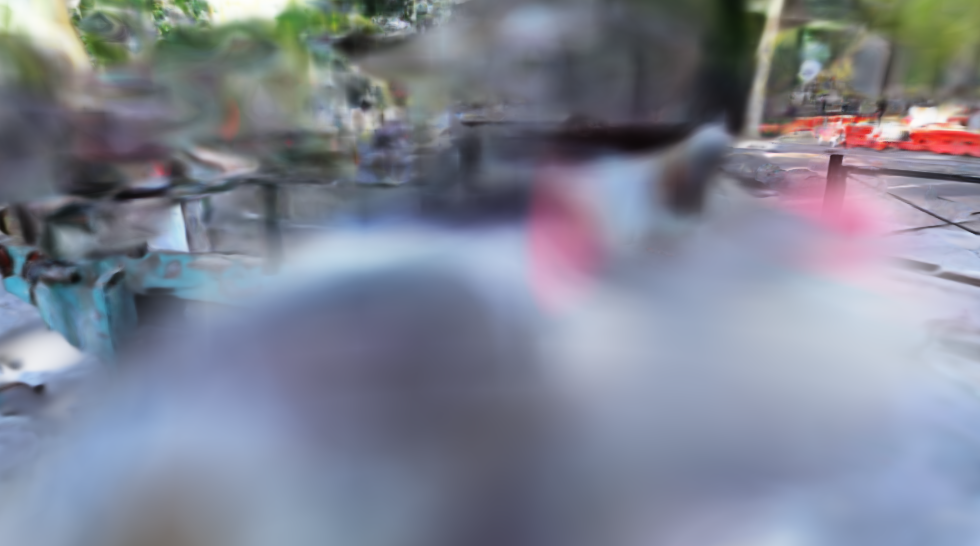}&
\includegraphics[width =0.66\columnwidth ]{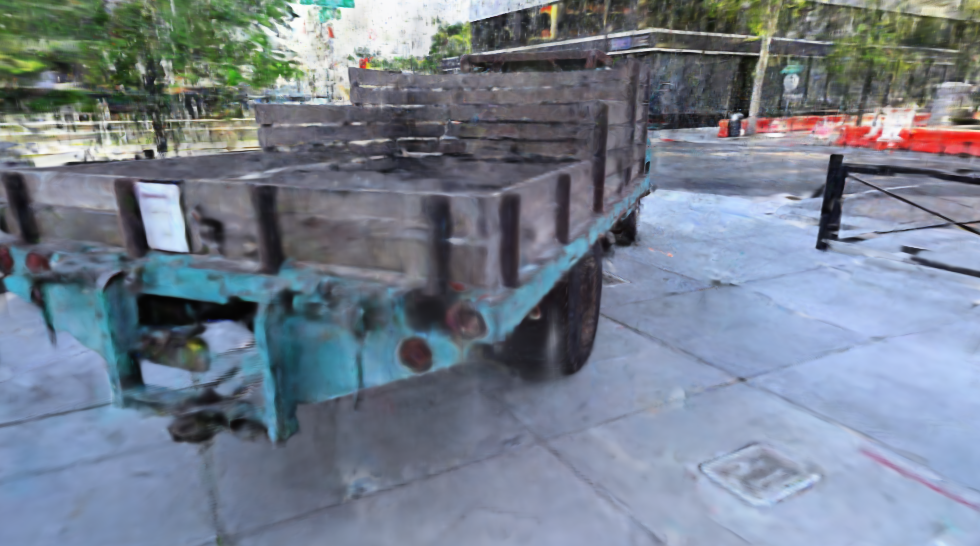}\\
\rotatebox[origin=l]{90}{\parbox{0.9in}{\centering Corrupted}} &
\includegraphics[width =0.66\columnwidth ]{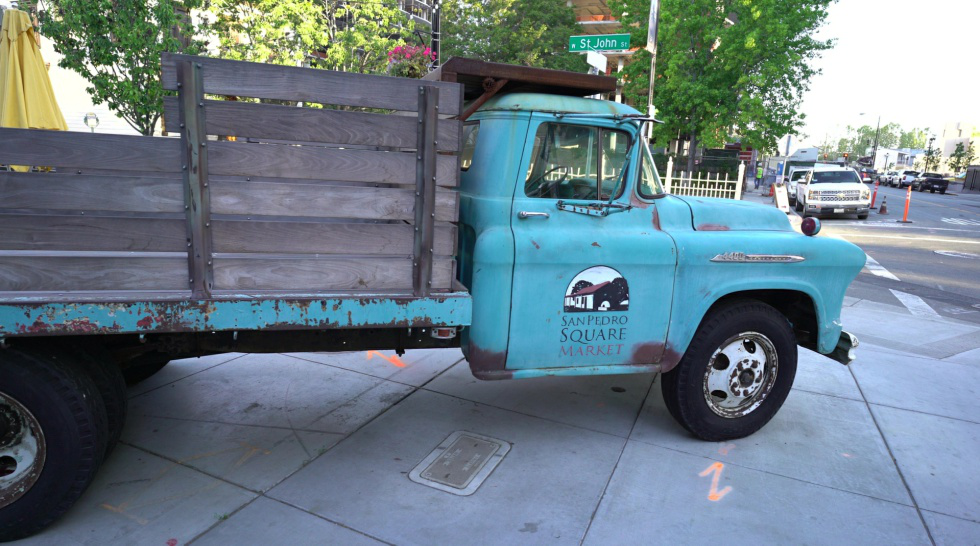}&
\includegraphics[width =0.66\columnwidth ]{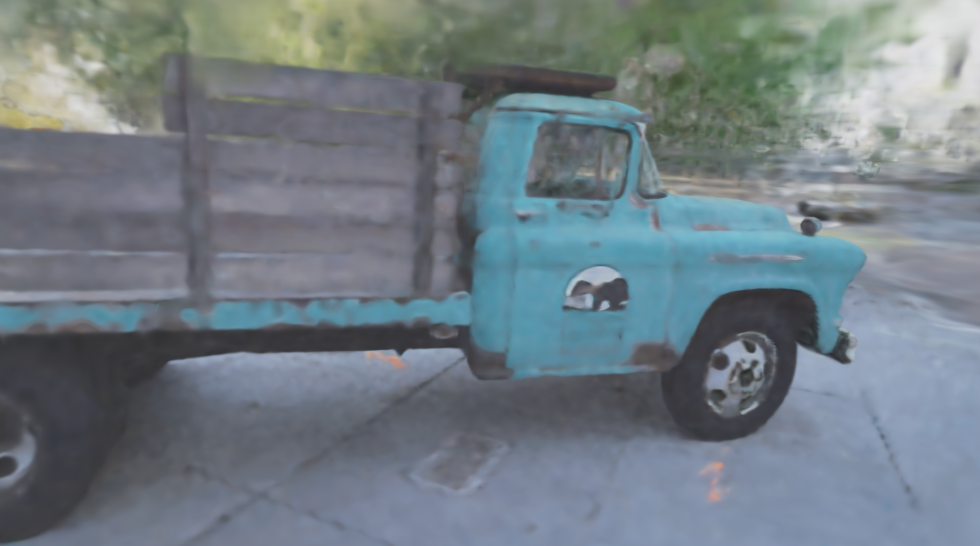}&
\includegraphics[width =0.66\columnwidth ]{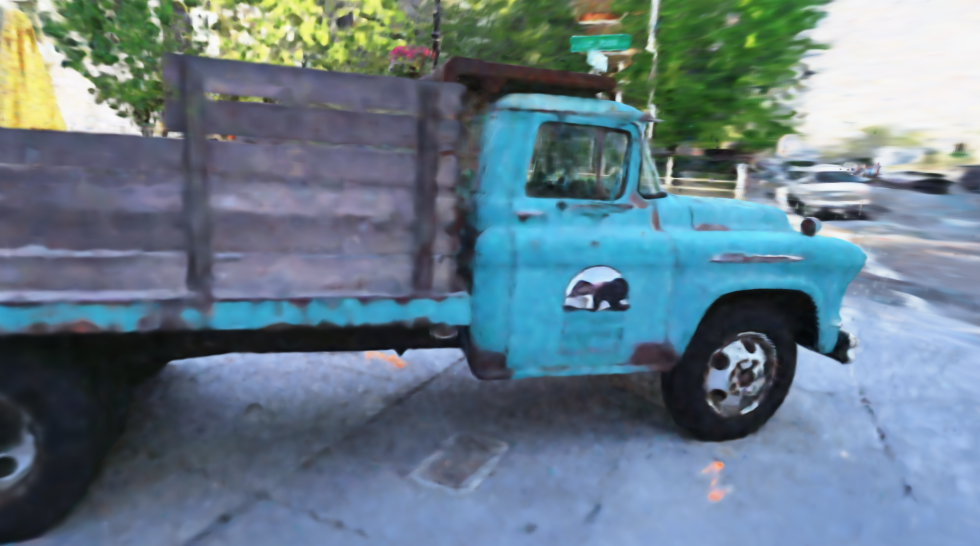} \\
\rotatebox[origin=l]{90}{\parbox{1.5in}{\centering Dynamic}} &
\includegraphics[width =0.66\columnwidth ]{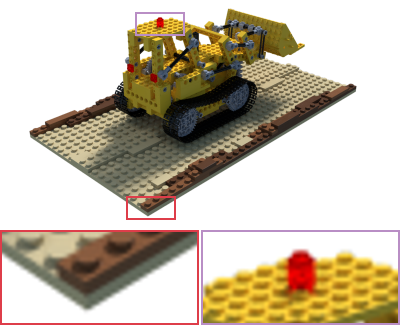}&
\includegraphics[width =0.66\columnwidth ]{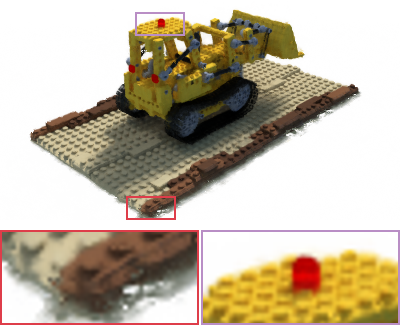}&
\includegraphics[width =0.66\columnwidth ]{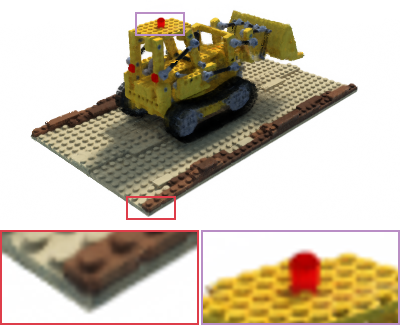} \\
\end{tabular}
\caption{Qualitative comparison of standard training and \textit{multiplex training} for neural radiance field. Top Row: Sparse Inputs. Middle Row: Corrupted Images. Bottom Row: Dynamic Scene.}
 \label{fig:dsc_nerfqualitative}
\end{figure*}

 
 We first choose the accelerated NeRF variants, DVGO, and TensoRF, as representatives of the NeRF family because training the accelerated NeRF methods is more environment-friendly and can significantly reduce the energy costs and carbon emissions of our work. 

Our quantitative results on eight Replica scenes and four T$\&$T scenes in Tables \ref{table:dvgoreplica}, \ref{table:tensorfreplica}, and \ref{table:nerfmeantntad} demonstrate that \textit{multiplex training} via S3IM remarkably improves all three common image quality metrics for the NeRF faimily, including vanila NeRF, DVGO, and TensoRF. Particularly, the mean improvement in PSNR over eight Replica scenes can be incredibly up to \textbf{16.43} and \textbf{24.75} for DVGO and TensoRF, respectively, over eight novel view synthesis tasks. The PSNR gain suggests that test MSE can decrease by 2-3 orders of magnitude due to S3IM. As PSNR directly depends on MSE, it is surprising to see the improvement in PSNR given the fact that S3IM distract the training objective from the original MSE loss. It means that S3IM significantly improve the generalization of NeRF.

Why is S3IM so important on Replica Dataset? We conjecture that this is because Replica contains more complex details but fewer training images, while T$\&$T Dataset contains less complex details but more training images. We will further verify this in the following experiments.

To the best of our knowledge, this is the most significant performance improvement along this line of research. We visualize the qualitative comparisons in Figure \ref{fig:nerfqualitative}. We leave the results of each T$\&$T scenes in Appendix C.


\begin{figure}[h]
\center
\subfigure{\includegraphics[width =0.325\columnwidth ]{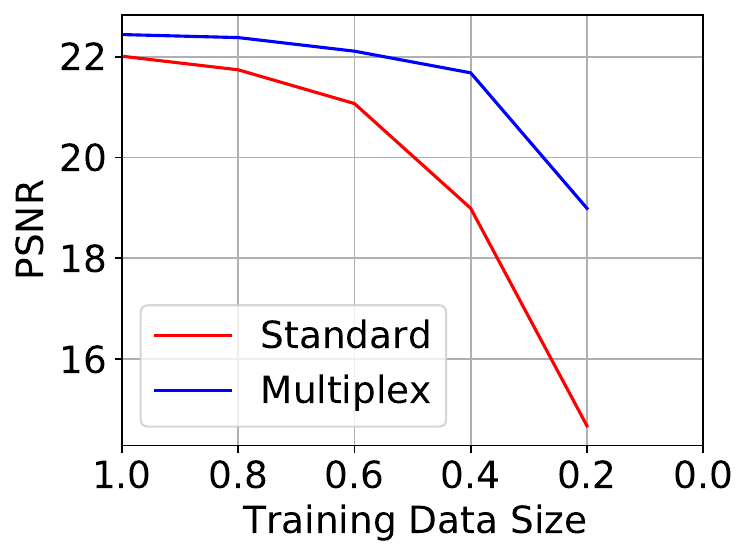}} 
\subfigure{\includegraphics[width =0.325\columnwidth ]{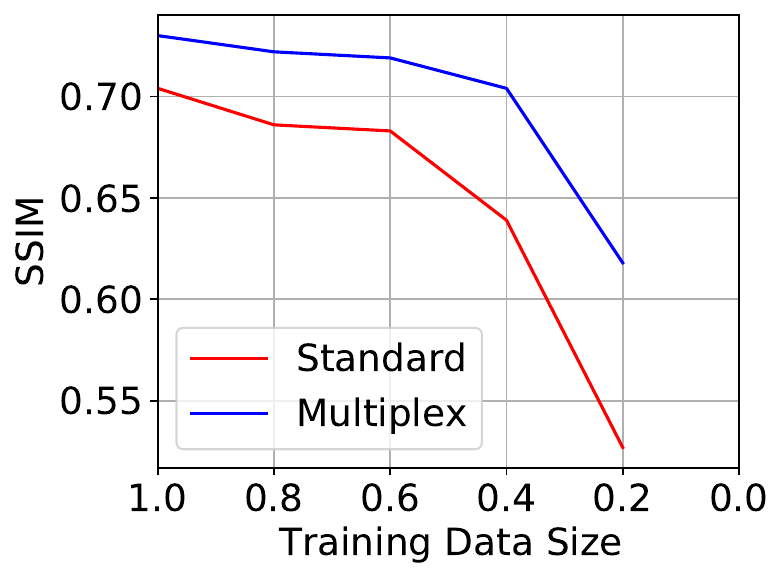}}
\subfigure{\includegraphics[width =0.325\columnwidth ]{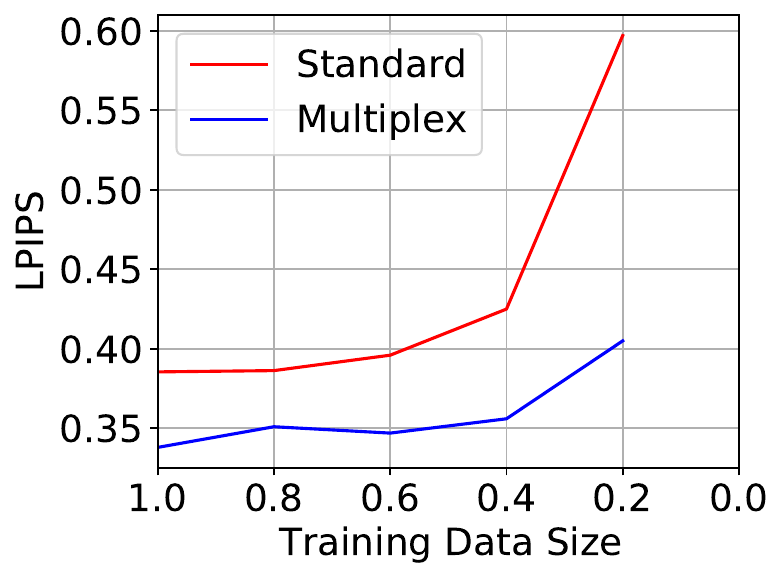}}
\caption{We plot the curves of PSNR, SSIM, and LPIPS with respect to the training data size, namely the portion of training samples kept from the original training dataset. The improvement of \textit{multiplex training} can be even more significant when the training data size decreases. Model: DVGO. Dataset: T$\&$T-Truck.}
 \label{fig:s3imsparse}
\end{figure}

\textbf{Few-shot Learning with sparse inputs.} Learning with sparse or very few examples is a hot topic in machine learning \cite{wang2020generalizing} as well as neural rendering. In practice, we may not be able to collect many images of a scene. Does S3IM work well even if we have very few images? To answer this question, we train DVGO with the sparse version of a simple truck scene from T$\&$T Intermediate, called Sparse Truck, where we randomly remove some training images.

 We visualize the qualitative comparisons in Figure \ref{fig:dsc_nerfqualitative}. The experimental results in Figure \ref{fig:s3imsparse} further suggest that the performance improvement of \textit{multiplex training} can be more significant when we have fewer training images. For example, when only $20\%$ of the original training dataset is available, the improvement in PSNR can be surprisingly up to 4.32 for the simple Truck scene of T$\&$T Intermediate. The more significant improvement with sparser inputs may explain why \textit{multiplex training} makes more significant improvements on Replica Dataset than T$\&$T Dataset.

\begin{figure}[h]
\center
\subfigure{\includegraphics[width =0.325\columnwidth ]{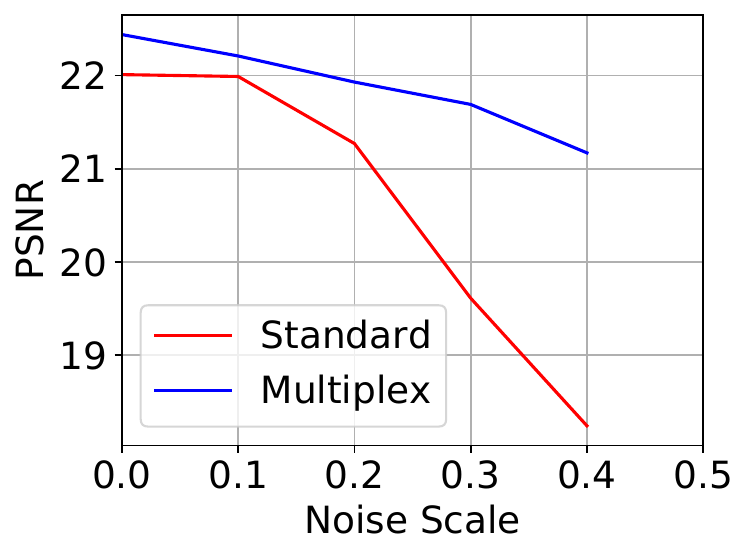}} 
\subfigure{\includegraphics[width =0.325\columnwidth ]{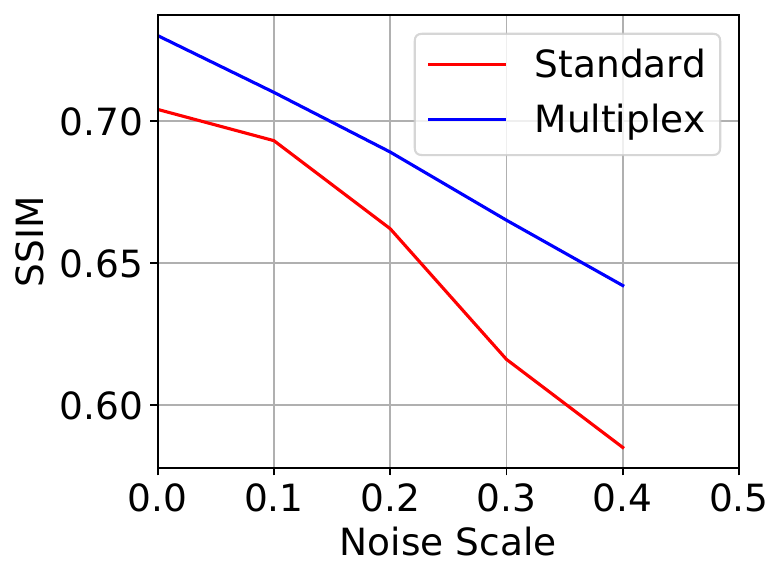}}
\subfigure{\includegraphics[width =0.325\columnwidth ]{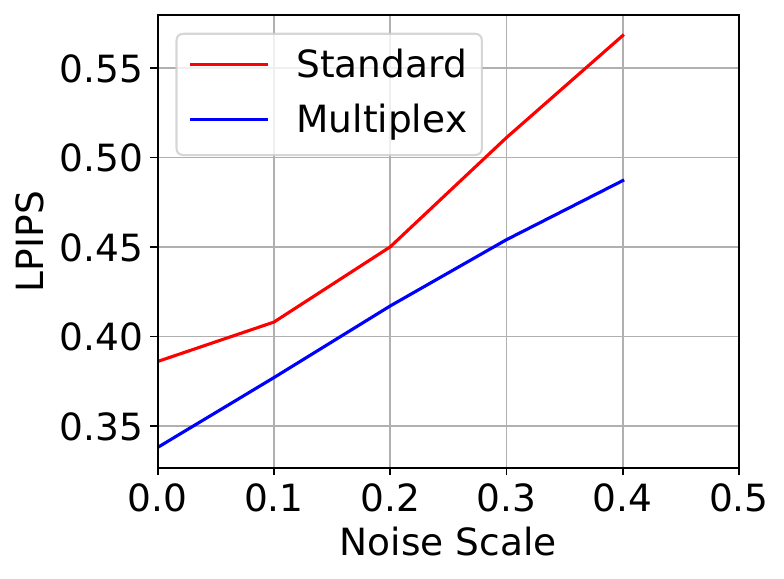}}
\caption{We plot the curves of PSNR, SSIM, and LPIPS with respect to the image noise scale $\std$. The improvement of \textit{multiplex training} can be even more significant when training image is corrupted by random noise. Model: DVGO. Dataset: T$\&$T-Truck.}
 \label{fig:s3imgaussian}
\end{figure}

\textbf{Robustness to image corruption.} While common benchmark datasets in NeRF studies are relatively clear, the training images in practice may be corrupted or noisy due to the realistic limitation of data collection. Gaussian image noise in digital images often arise during acquisition due to the inherent noise in the sensors \cite{boyat2015review}. Robustness to data corruption and noise memorization can be an important performance metric for neural networks \cite{xie2021artificial} in weakly-supervised learning but unfortunately overlooked by most existing NeRF studies. 

We again use the Truck scene to make an image-corrupted dataset, called Corrupted Truck, where we inject Gaussian noise with the standard deviation as $\std$ into original Truck images (each RGB value lies in $[0,1]$) and obtain the corrupted version. We visualize the qualitative comparisons in Figure \ref{fig:dsc_nerfqualitative}. The experimental results in Figure \ref{fig:s3imgaussian} show that \textit{multiplex training} via S3IM can significantly improve the robustness to image corruption.

\begin{table}[h]
\caption{Quantitative results of NeRF methods on dynamic scenes, Mutant and LEGO \cite{pumarola2021d}. Model: D-NeRF.}
\vspace{-0.2in}
\label{table:dnerf}
\begin{center}
\begin{small}
\resizebox{0.48\textwidth}{!}{%
\begin{tabular}{c|c|ccc}
\toprule
Scene & Training &  PSNR($\uparrow$) & SSIM($\uparrow$) & LPIPS($\downarrow$)\\
\midrule 
\multirow{2}{*}{Lego}  & Standard & 21.29 & 0.821 & 0.0634 \\
  & Multiplex & \textbf{23.36} & \textbf{0.900} & \textbf{0.0482}  \\   
\midrule 
\multirow{2}{*}{Mutant}  & Standard & 32.14 & 0.972 & 0.0181 \\
 & Multiplex & \textbf{32.76} & \textbf{0.976} & \textbf{0.0151} \\  
\bottomrule
\end{tabular}
}
\end{small}
\end{center}
\end{table}

\textbf{Dynamic scene synthesis.} Rendering novel photo-realistic views of dynamic scenes is a more difficult task than neural rendering of static scenes. A NeRF variant, D-NeRF, has been specifically designed to render dynamic scenes. We also study how \textit{multiplex training} via S3IM improves D-NeRF on two representative dynamic scenes, Lego and Mutant \cite{pumarola2021d}. We visualize the qualitative comparisons in Figure \ref{fig:dsc_nerfqualitative}. The results of Table \ref{table:dnerf} and Figure \ref{fig:dsc_nerfqualitative} both suggest that \textit{multiplex training} via S3IM also significantly improve the performance of D-NeRF for dynamic scenes.

\textbf{Regularization methods for monocular video.} Moreover, we also show that S3IM can further improve other regularized methods specifically designed for monocular video, such as Dynamic NeRF \cite{gao2021dynamic}. Monocular video only contains sparse and continuous training views. We reproduce the original Dynamic NeRF's quantitative result on the monocular video (Balloon1) and compare it with S3IM-enhanced results in Table \ref{table:dynamicnerf}.

\begin{table}[h]
\caption{Quantitative results of Dynamic NeRF methods on monocular video \cite{gao2021dynamic}. Model: Dynamic NeRF.}
\label{table:dynamicnerf}
\begin{center}
\begin{small}
\resizebox{0.45\textwidth}{!}{%
\begin{tabular}{c|c|ccc}
\toprule
Scene & Training &  PSNR($\uparrow$) & SSIM($\uparrow$) & LPIPS($\downarrow$)\\
\midrule 
\multirow{2}{*}{Balloon1}  & Standard & 19.80 & 0.693 & 0.210 \\
  & Multiplex & \textbf{22.55} & \textbf{0.767} & \textbf{0.108}  \\   
\bottomrule
\end{tabular}
}
\end{small}
\end{center}
\end{table}

\begin{table*}[t]
\caption{Quantitative results of neural surface reconstruction on Replica Dataset \cite{replica19arxiv}. Model: NeuS.}
\label{table:neusreplica}
\begin{center}
\begin{small}
\resizebox{0.95\textwidth}{!}{%
\begin{tabular}{c|c|cccccccccc}
\toprule
Scene & Training & PSNR($\uparrow$) & SSIM($\uparrow$) & LPIPS($\downarrow$) & Chamfer-$L_{1}$($\downarrow$) & Accuracy($\downarrow$) & Completeness($\downarrow$) & Precision($\uparrow$) & Recall($\uparrow$) & F-score($\uparrow$) & Normal C.($\uparrow$) \\
\midrule 
\multirow{2}{*}{Scene 1}  & Standard & 28.52 & 0.843 & 0.148  & 18.13 & 11.60 & 24.65 & 29.84 & 17.79 & 22.29 & 72.14   \\
 & Multiplex & \textbf{30.93} & \textbf{0.876} & \textbf{0.100} & \textbf{6.264} & \textbf{6.580} & \textbf{5.948} & \textbf{58.31} & \textbf{81.00} & \textbf{59.99} & \textbf{80.58}  \\
\midrule 
 \multirow{2}{*}{Scene 2} & Standard & 29.15 & 0.834 & 0.179  & 19.00 & 12.73 & 25.27 & 28.32 & 16.29 & 20.69 &  \textbf{75.16}  \\
 & Multiplex & \textbf{31.06} & \textbf{0.870} & \textbf{0.135} & \textbf{7.475} & \textbf{8.506} & \textbf{6.425} & \textbf{53.48} & \textbf{61.25} & \textbf{57.10} & 72.84  \\ \midrule 
 \multirow{2}{*}{Scene 3} & Standard & 28.44 & 0.877 & 0.150  & 40.34 & 33.80 & 46.89 & 9.47 & 5.68 & 7.10 & 67.69   \\
 & Multiplex & \textbf{33.97} & \textbf{0.933} & \textbf{0.0372} & \textbf{6.188}  & \textbf{6.462} & \textbf{5.914} & \textbf{57.66} & \textbf{59.95} & \textbf{58.72} & \textbf{75.96}   \\ 
 \midrule 
\multirow{2}{*}{Scene 4}  & Standard & 31.84 & 0.874 & 0.152  & 18.90 & 13.61 & 24.19 & 22.25 & 12.52 & 16.02 & 71.89   \\
 & Multiplex & \textbf{37.28} & \textbf{0.940} & \textbf{0.0596} & \textbf{7.397} & \textbf{8.223} & \textbf{6.570} & \textbf{54.60} & \textbf{70.68} & \textbf{61.60} & \textbf{72.97}   \\   
  \midrule 
 \multirow{2}{*}{Scene 5} & Standard & 33.78 & 0.897 & 0.121  & 76.33 & 11.83 & 140.83 & 33.09 & 0.60 & 1.18 & 50.05   \\
 & Multiplex & \textbf{36.32} & \textbf{0.924} & \textbf{0.071} & \textbf{34.40} & \textbf{31.43} & \textbf{37.38} & \textbf{14.89} & \textbf{9.56} & \textbf{11.64} & \textbf{55.60}   \\ 
   \midrule 
 \multirow{2}{*}{Scene 6} & Standard & 27.82 & 0.882 & 0.141  & 15.63 & 11.71 & 19.54 & 30.74 & 21.53 & 25.32 & 69.33  \\
 & Multiplex & \textbf{31.18} & \textbf{0.914} & \textbf{0.0995} & \textbf{6.980} & \textbf{7.211} & \textbf{6.748} & \textbf{60.08} & \textbf{59.88} & \textbf{59.98} & \textbf{73.62}   \\ 
\midrule 
 \multirow{2}{*}{Scene 7} & Standard & 28.80 & 0.898 & 0.114 & 11.45 & 8.88 & 14.02 & 43.23 & 33.39 & 37.68 & \textbf{78.53}   \\
 & Multiplex & \textbf{30.99} & \textbf{0.917} & \textbf{0.0837} & \textbf{7.824} & \textbf{7.845} & \textbf{7.803} & \textbf{51.38} & \textbf{51.23} & \textbf{51.31} & 78.30  \\
\midrule 
 \multirow{2}{*}{Scene 8} & Standard & 28.29 & 0.908 & 0.130  & 30.82 & 25.71 & 35.92 & 15.34 & 8.73 & 11.13 & 68.86   \\
 & Multiplex & \textbf{34.89} & \textbf{0.954} & \textbf{0.0539} & \textbf{6.136} & \textbf{5.839} & \textbf{6.434} & \textbf{61.62} & \textbf{62.56} & \textbf{62.08} & \textbf{79.68}  \\
  \midrule 
 \multirow{2}{*}{Mean (8 scenes)} & Standard & 29.58 & 0.877 & 0.142  & 28.83 & 16.23 & 41.41 & 26.54 & 14.57 & 17.68 & 69.21  \\
 & Multiplex & \textbf{33.33} & \textbf{0.916} & \textbf{0.0799} & \textbf{10.33} & \textbf{10.26} & \textbf{10.40} & \textbf{51.50} & \textbf{57.01} & \textbf{52.80} & \textbf{73.69}  \\
\bottomrule
\end{tabular}
}
\end{small}
\end{center}
\end{table*}

\subsection{Surface Reconstruction Experiments}

\begin{table}[h]
\caption{Quantitative results of NeuS on T$\&$T.}
\label{table:neustntad}
\begin{center}
\begin{small}
\resizebox{0.45\textwidth}{!}{%
\begin{tabular}{c|c|ccc}
\toprule
 Scene & Training & PSNR($\uparrow$) & SSIM($\uparrow$) & LPIPS($\downarrow$) \\
\midrule 
\multirow{2}{*}{Scene 1}  & Standard & 20.73 & 0.636 & 0.393 \\
  & Multiplex & \textbf{22.01} & \textbf{0.681} & \textbf{0.311} \\   
\midrule 
\multirow{2}{*}{Scene 2}  & Standard  & 21.26 & 0.739 & 0.434  \\
  & Multiplex & \textbf{23.23} & \textbf{0.812} & \textbf{0.270} \\
\midrule 
\multirow{2}{*}{Scene 3}  & Standard & 17.58 & 0.551 & 0.428 \\
  & Multiplex & \textbf{19.07} & \textbf{0.609} & \textbf{0.344} \\
\midrule 
\multirow{2}{*}{Scene 4}  & Standard & 20.32 & 0.554 & 0.398 \\
  & Multiplex & \textbf{22.90} & \textbf{0.660} & \textbf{0.301}  \\
\midrule 
\multirow{2}{*}{Mean}  & Standard & 19.97 & 0.620 & 0.413 \\
  & Multiplex & \textbf{21.55} & \textbf{0.691} & \textbf{0.307}  \\
\bottomrule
\end{tabular}
}
\end{small}
\end{center}
\end{table}


Reconstructing surfaces from images is also a fundamental problem in computer vision. Recent neural surface reconstruction methods \cite{yariv2020multiview,yariv2021volume,oechsle2021unisurf,wang2021neus} belong to another kind of neural field method orthogonal to NeRF. To evaluate the universal effectiveness of the proposed method, we empirically study S3IM for a classical neural surface reconstruction method, called NeuS, which can render both RGB images and surface information. NeuS employs the same training data as NeRF without ground-truth surface information.

We use Replica Dataset and T$\&$T Advanced as two benchmark datasets for surface reconstruction. Replica Dataset has the ground-truth surface information in test data, Advanced Scenes of T$\&$T Dataset has no available ground-truth surface information. We add a group quality metrics which can measure rendering quality of surface information and geometric information when ground-truth surface and geometric information is available. These surface quality metrics, especially such as Chamfer $L_1$ Distance and F-score, are widely used for evaluating surface reconstruction.

\begin{figure}[h]
\center
\renewcommand*{\arraystretch}{0}
\begin{tabular}{@{}c@{}c@{}c@{}c@{}}
 & Ground Truth & Standard & Multiplex(Ours)\\
\rotatebox[origin=l]{90}{\parbox{0.9in}{\centering RGB}} &
\includegraphics[width =0.32\columnwidth ]{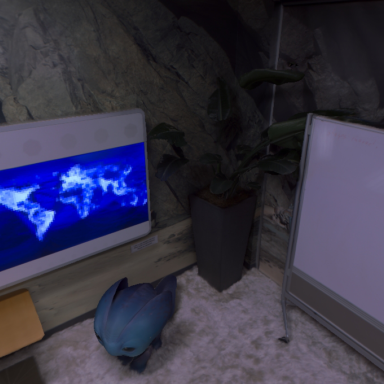}&
\includegraphics[width =0.32\columnwidth ]{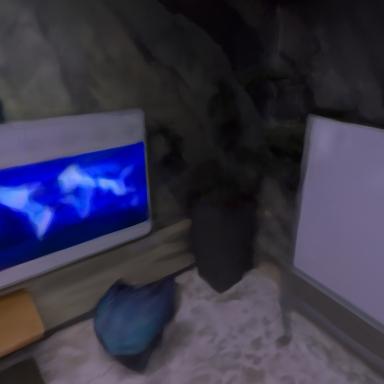}&
\includegraphics[width =0.32\columnwidth ]{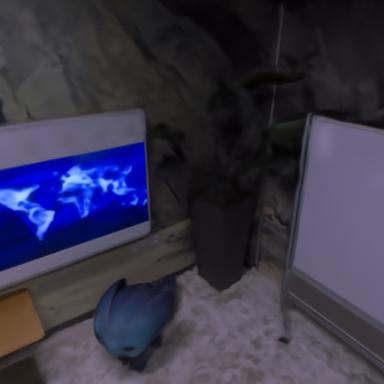}\\
\rotatebox[origin=l]{90}{\parbox{0.8in}{\centering Depth}} &
\includegraphics[width =0.32\columnwidth ]{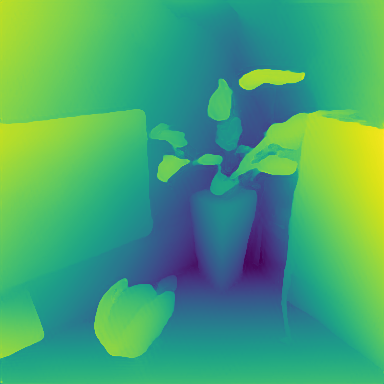}&
\includegraphics[width =0.32\columnwidth ]{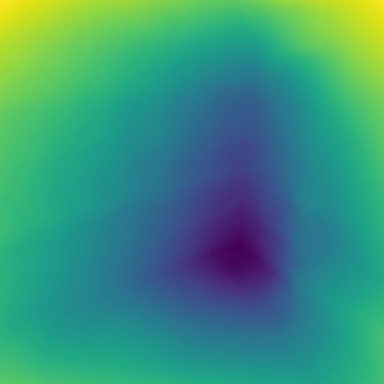}&
\includegraphics[width =0.32\columnwidth ]{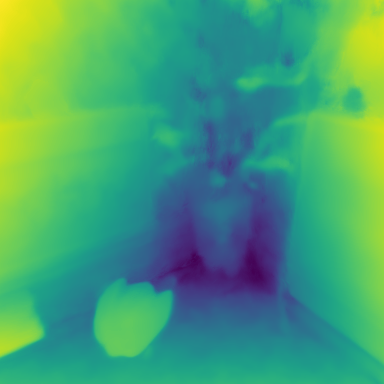}\\
\end{tabular}
\caption{RGB rendering and depth rendering of standard training and \textit{multiplex training} via S3IM for neural surface representation. Model: NeuS. Dataset: Replica Scene 1 (Room 0).}
 \label{fig:neusqualitative}
\end{figure}

Our qualitative results in Figure \ref{fig:neusqualitative} show that S3IM improve both RGB rendering and depth rendering. Our quantitative results in Tables \ref{table:neusreplica} and \ref{table:neustntad} suggest that \textit{multiplex training} via S3IM can significantly improve neural surface reconstruction methods in terms of all three image quality metrics and all seven surface quality metrics. For example, in terms if two very popular surface quality metrics, we obtain \textbf{a $\mathbf{64\%}$ Chamfer $\mathbf{L_{1}}$ distance reduction and a $\mathbf{198\%}$ F-score gain over eight surface reconstruction tasks}. Most surface quality metrics have been improved by more than 10 points. This again demonstrate the unreasonable effectiveness and universality of S3IM in neural field methods.

\subsection{Discussion}

\begin{figure}[h]
\center
\subfigure{\includegraphics[width =0.325\columnwidth ]{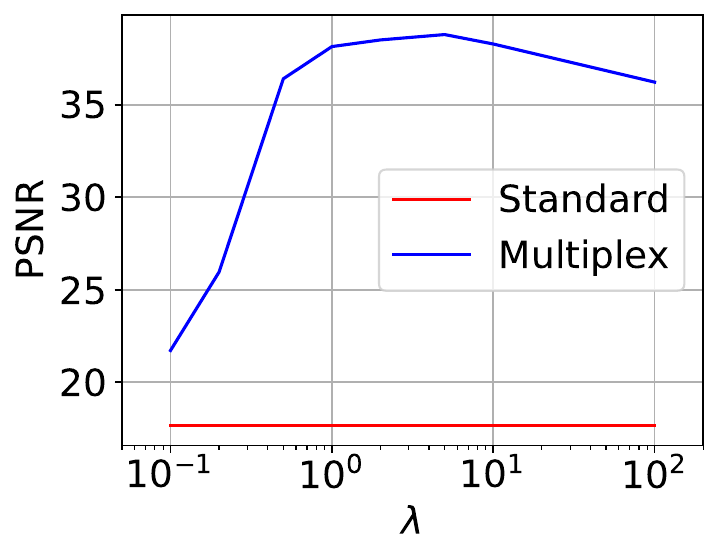}} 
\subfigure{\includegraphics[width =0.325\columnwidth ]{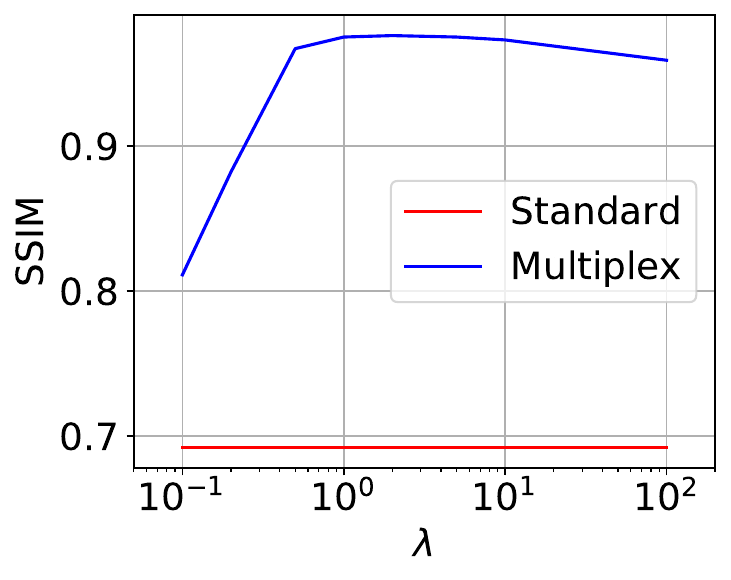}}
\subfigure{\includegraphics[width =0.325\columnwidth ]{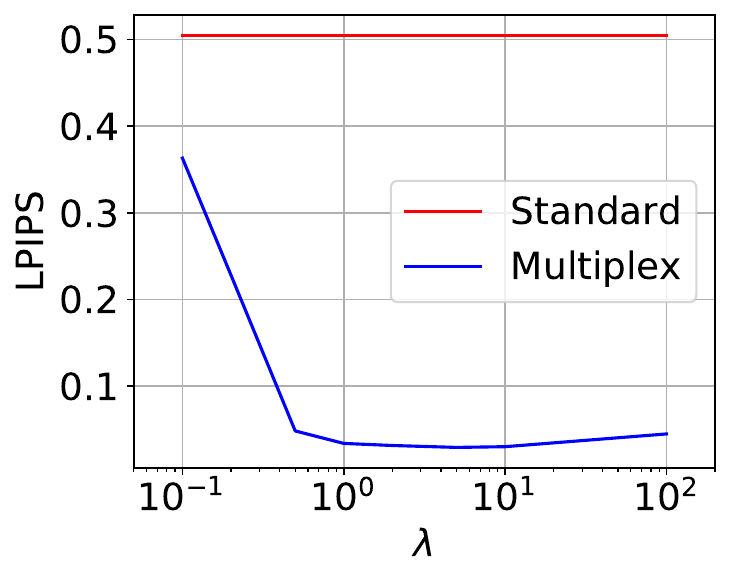}}
\vspace{-0.2in}
\caption{We plot the curves of PSNR, SSIM, and LPIPS with respect to the hyperparameter $\lambda$. S3IM is robust to a wide range of $\lambda$. Model: DVGO. Dataset: Replica Scene 4.}
 \label{fig:s3imlambda}
\end{figure}

\textbf{S3IM hyperparameters $\{\lambda, M, K\}$.} We plot the curves of PSNR, SSIM, and LPIPS with respect to the hyperparameter $\lambda$ in Figure \ref{fig:s3imlambda}. The improvement of S3IM is robust to a wide range of $\lambda$ choices. We also study how the improvement depends on the hyperparameter $M$ and report that $M=1$ can also make significant improvements, while the default value is $M=10$, shown in Table \ref{table:robusttimeM}. 

Moreover, we also empirically verify that choosing a relatively small kernel size for S3IM provide robust statistics as we discuss. We empirically evaluated S3IM with the kernel size $K$ and stride size $S$ as 4, 16, and 64, respectively. Note that the computational complexity is approximately invariant to the kernel size, when we choose $K=S$. The setting $K=S$ can provide fair comparisons. In Table \ref{table:kernelS3IM}, we clearly observe that increasing the size $K$ can significantly degrade the performance.

\begin{table}[h]
\caption{Quantitative results and the training time (A100 GPU hours) with respect to the hyperparameter $M$. Dataset: Replica}
\vspace{-0.2in}
\label{table:robusttimeM}
\begin{center}
\begin{small}
\resizebox{0.48\textwidth}{!}{%
\begin{tabular}{c|c|c|c|ccc|c}
\toprule
 Scene & Model & $M$ & Training & PSNR($\uparrow$) & SSIM($\uparrow$) & LPIPS($\downarrow$) & Training Time\\
 \midrule 
\multirow{3}{*}{Room 0} & \multirow{3}{*}{TensoRF} & 
  0  & Standard & 12.03 & 0.464 & 0.773 & 0.369 \\  
& & 1  & Multiplex &  36.65 & 0.954 & 0.0387  & 0.374  \\ 
& & 10  & Multiplex &  37.15 & 0.958 & 0.0335  & 0.432  \\ 
 \midrule
\multirow{3}{*}{Office 0}  & \multirow{3}{*}{NeuS} & 0  & Standard & 31.84 & 0.874 & 0.152  & 2.95 \\ 
 & & 1 & Multiplex & 37.02 & 0.937 & 0.0666 & 2.95   \\ 
 & & 10 & Multiplex & 37.28 & 0.940 & 0.0596 & 2.98   \\  
\bottomrule
\end{tabular}
}
\end{small}
\end{center}
\end{table}

\begin{table}[h]
\caption{Quantitative results of S3IM with various kernel sizes. Model: TensoRF. Dataset: Replica Scene 1 (Room 0).}
\vspace{-0.2in}
\label{table:kernelS3IM}
\begin{center}
\begin{small}
\resizebox{0.48\textwidth}{!}{%
\begin{tabular}{c|c|ccc}
\toprule
Kernel Size & Training & PSNR($\uparrow$) & SSIM($\uparrow$) & LPIPS($\downarrow$) \\
-  & Standard & 12.13 & 0.468 & 0.7786  \\
4  & Multiplex & \textbf{37.15} & \textbf{0.958} & \textbf{0.0335} \\
16  & Multiplex & 36.79 & 0.954 & 0.0375 \\
64  & Multiplex & 12.10 & 0.4989 & 0.8124 \\ 
\bottomrule
\end{tabular}}
\end{small}
\end{center}
\end{table}

\textbf{Computational costs.} We also present the training time (GPU hours) of standard training and multiplex training with respect to $M$ in Table \ref{table:robusttimeM}. The results show that the extra computational cost of S3IM is very limited compared with the significant quality improvement. For example, the extra computational costs are only $8\%$ for TensoRF and $1\%$ for NeuS with $M=10$; only $1\%$ for TensoRF and nearly free for NeuS with $M=1$.

\begin{table}[h]
\caption{Ablation study of nonlocal S3IM and local SSIM.}
\vspace{-0.1in}
\label{table:ablationnerf}
\begin{center}
\begin{small}
\resizebox{0.48\textwidth}{!}{%
\begin{tabular}{c|c|c|ccc}
\toprule
Scene & Model & Training & PSNR($\uparrow$) & SSIM($\uparrow$) & LPIPS($\downarrow$)\\
 \midrule 
\multirow{3}{*}{Truck}  &\multirow{3}{*}{DVGO}  & MSE & 22.01 & 0.704 & 0.386  \\
 & & MSE + SSIM & 15.90 & 0.569 & 0.566 \\
 & & MSE + S3IM (Ours) & \textbf{22.44} & \textbf{0.730} & \textbf{0.338} \\ 
\bottomrule
\end{tabular}
}
\end{small}
\end{center}
\end{table}

\textbf{S3IM verus SSIM.} 
We conduct ablation study on the proposed S3IM over stochastic patches and the conventional image quality metric SSIM over local patches. The patch sizes of S3IM and SSIM are both $64\times64$, while the kernel/stride sizes are both 4.. We present the ablation study of stochastic-patch S3IM and local-patch SSIM in Table \ref{table:ablationnerf}. The results suggest that S3IM with the non-local and stochastic structural information is very helpful, while SSIM with the local structural information is marginal and sometimes even harmful due to lack of stochasticity. The ablation study again verifies the novel contribution of our method.

\textbf{Future Work} There are at least three promising future research directions. First, we may directly introduce S3IM into non-RGB losses, such as depth losses. Second, we can develop better multiplex losses than S3IM for other machine learning tasks, including Graph Neural Networks and Physics-Informed Neural Network \cite{raissi2019physics}, as long as the point-wise losses are optimized in these tasks. Third, it will be very valuable to theoretically understand minima's flatness \cite{zhang2017understanding,xie2021diffusion} and generalization learned by S3IM.

\section{Conclusion}
\label{sec:conclusion}

Recently neural fields have achieved great empirical success and are receiving considerable attention in computer vision and computer graphics. However, the current training paradigm usually uses only point-wise supervision information and overlooks the rich structural information contained in the group of pixels. In this work, we proposed S3IM to extend the performance limit of neural fields by exploiting the nonlocal structural information of groups of pixels. We have demonstrated the unreasonable and robust effectiveness of S3IM for all employed models and scenes in terms of 10 quality metrics, while the extra costs are nearly free. Our extensive experiments strongly support important values of S3IM and nonlocal information. We believe that S3IM will serve as a default method for training neural fields in future. 

{\small
\bibliographystyle{ieee_fullname}
\bibliography{nerf}
}

\newpage

\appendix

\section{Experimental Settings and Details}
\label{sec:expdetail}

In this section, we present the experimental settings and details for reproducing the results. The main principle of our experimental setting is to fairly compare \textit{multiplex training} and standard training for NeRF and the variants. Our experimental settings follows original papers to produce the baselines, unless we specify otherwise.

\textbf{S3IM Setting.} We always choose the kernel size $K=4$, the stride size $S=4$, and the number of stochastic patches $M=10$ without fine-tuning. We fine-tune the S3IM loss weight $\lambda$ from $\{0.05,0.1,0.2,0.5,1, 2, 5\}$ for the NeRF family and $\{1,2, 5, 10, 20, 50, 100\}$ for the NeuS family. 

\subsection{Models and Optimization}

\textbf{DVGO Setting} We employ the sourcecode of DVGO (Version 2) in the original project \cite{sun2022direct} without modifying training hyperparameters. So we train DVGO via Adam \cite{kingma2014adam} with the batch size $B=8192$. The learning rate of density voxel grids and color/feature voxel grids is 0.1, and the learning rate of the RGB net (MLP) is 0.001. The total number of iterations is 5000. We multiply the learning rate by 0.1 per 1000 iterations. 

\textbf{TensoRF Setting} We employ the sourcecode of the original project \cite{chen2022tensorf} without modifying training hyperparameters. The total number of iterations is 30000. The batch size is 4096. The initial number of voxels is $128^{3}$, while the final number of voxels is $300^{3}$. The upsamping iterations for voxels are 2000,3000,4000,5500 and 7000, respectively. Adam with $\beta_{1}= 0.9$ and $\beta_{2}= 0.99$ is used.

\textbf{NeRF Setting} We employ a popular open-source implementation \cite{lin2020nerfpytorch} of the original NeRF. Again, we follow its defaulted training setting. The learning rate is $0.0005$, and the learning rate scheduler is $0.1^{iters/500000}$. 

\textbf{D-NeRF Setting} We directly employ the sourcecode of D-NeRF in the original project \cite{pumarola2021d}. We only slightly change the original batch size $B=500$ to $B=512$ for generating the squared stochastic patch. The learning rate is 0.0005. The total number of iterations is 800k. The learning rate decay follows the original paper. 

\textbf{NeuS Setting} We employ the NeuS implementation of SDFStudio \cite{Yu2022SDFStudio} and follow its default hyperparameters. The difference of the hyperparameters between SDFStudio and the original paper \cite{wang2021neus} is that SDFStudio trains 100k iterations, while the original paper trains 300k iterations. 

\subsection{Datasets}

\textbf{Replica Dataset} Replica Dataset has no splitted training dataset and test dataset. In the experiments on Replica, if one image index is divisible by 10, we move the image to the test dataset; if not, we move the image to the training dataset. Thus, we have $90\%$ images for training and $10\%$ images for evaluation.

\textbf{T$\&$T Dataset Advanced} T$\&$T Dataset Advanced has no splitted training data and test data. We follow the original splitted way in the standard setting. In the experiments on Replica, if one image index is divisible by 10, we move the image to the test T$\&$T Dataset Advanced; if not, we move the image to the training dataset. Similarly, we again have $90\%$ images for training and $10\%$ images for evaluation.

\textbf{T$\&$T Dataset Intermediate} T$\&$T Dataset has splitted training data and test data. We follow the original splitted way in the standard setting. In the experiments of sparse inputs, we randomly remove the training images. In the experiments of corrupted images, we inject Gaussian noise with the scale $\std$ into RGB values of the training images, and clip the corrupted RGB values into $[0,1]$.

\textbf{Dynamic Scenes: LEGO and Mutant} The two dynamics scenes are used in the original D-NeRF paper. We use them in the same way without any modification.

\section{Image Quality Metrics}
\label{sec:suppmetrics}

As we mentioned above, PSNR and SSIM are two most popular image quality metrics. Beyond them, we have also seen other useful metrics. Multiscale SSIM \cite{wang2003multiscale} is a variant of SSIM, which incorporates image details at different resolutions. However, Multiscale SSIM still can only capture local structural information carried by nearby pixels.

Deep features learned by DNNs has unreasonable effectiveness as a perceptual metric for measuring the similarity between two sets of perceptual features \cite{zhang2018unreasonable}. Thus, the Learned Perceptual Image Patch Similarity (LPIPS) metric \cite{zhang2018unreasonable} serves as the third rendering quality metric in NeRF and related studies, because it agrees surprisingly well with humans. The Fréchet inception distance (FID) score \cite{heusel2017gans} is another metric which measure the distance-based similarity of perceptive features, but it is more widely used for evaluating generative models, such as Generative Adversarial Networks \cite{creswell2018generative,goodfellow2020generative} (GAN) and Diffusion Models \cite{sohl2015deep,song2019generative,ho2020denoising}. Both LPIPS and FID metrics inevitably have certain stochasticity, because deep features are learned from stochastic training of DNNs. However, the stochasticity does not affect the usefulness and popularity of LPIPS and FID as image quality metrics.

S3IM can also be considered as a image quality metric, while we only use S3IM as a differentiable training objective in this paper. More specifically, S3IM measures the structural similarity of two paired sets of pixels(/signals), which may or may not form images. By analyzing the stochastic patch consists of  random pixels, S3IM can capture non-local structural information carried by nearby/distant pixels. S3IM also inevitably have certain stochasticity like LPIPS and FID, while S3IM does not depend on deep features given by DNNs.

SSIM is a well-known quality metric that can capture local structural similarity between images or patches. SSIM is considered to be correlated with the quality perception of the human visual system well and is widely used for evaluating NeRF \cite{wang2004image,hore2010image}. Suppose $\bm{a} = \{a_{i} | i=1,2,3, \dots, n\}$ and $\bm{b} = \{b_{i} | i=1,2,3, \dots, n\}$ to be
two discrete non-negative signals paired with each other (e.g. two image patches extracted from the same spatial location from paired images). We denote the mean intensity of a signal as $\mu$ (e.g. $\mu_{a} = \frac{1}{n}\sum_{i=1}^{n} a_{i}$), the standard deviation of a signal as $\sigma^{2}$ (e.g. $\sigma_{a}^{2} =  \frac{1}{n-1} \sum_{i=1}^{n} (a_i - \mu_a)^{2} $ ), and the covariance between two signals as $\sigma_{ab}^{2}$ (e.g. $\sigma_{ab}^{2} = \frac{1}{n-1} \sum_{i=1}^{n} (a_{i} - \mu_{a})(b_{i} - \mu_{b})$).

SSIM is expressed by the combination of three terms which are the luminance, contrast, and structure comparison metrics:
\begin{align}
\label{eq:ssim}
\SSIM(a, b) = l(\bm{a}, \bm{b}) c(\bm{a}, \bm{b}) s(\bm{a}, \bm{b}).
\end{align}
The luminance $l(\bm{a}, \bm{b})$, contrast $c(\bm{a}, \bm{b})$, and structure comparison $s(\bm{a}, \bm{b})$ are, respectively, written as
\begin{align}
\label{eq:lcs}
l(\bm{a}, \bm{b}) =& \frac{2\mu_{a}\mu_{b} + C_{1}}{\mu_{a}^{2} + \mu_{b}^{2} + C_{1}},\\
c(\bm{a}, \bm{b}) =& \frac{2\sigma_{a}\sigma_{b} + C_{2}}{\sigma_{a}^{2} + \sigma_{b}^{2} + C_{2}},\\
s(\bm{a}, \bm{b}) =& \frac{\sigma_{ab} + C_{3}}{\sigma_{a}\sigma_{b}  + C_{3}},
\end{align}
where $C_1$, $C_2$, and $C_3$ are small constants given by 
\begin{align}
\label{eq:constants}
C_1 = (K_{1}L)^{2}\text{, }C_2 = (K_{2}L)^{2}\text{, and }C_3= C_2 / 2.
\end{align}
Following the common setting \cite{wang2004image,mildenhall2021nerf}, we set $K_{1}=0.01$ and $K_{2}=0.03$ in this paper. The data range $L$ is $1$ for pixel RGB values. The range of SSIM lies in $[-1,1]$.

In practice of image quality assessment, people usually apply the SSIM index locally rather than globally. The local statistics $\mu_{a}$, $\sigma_{a}$, and $\sigma_{ab}$ are computed within a local $K\times K$ kernel window, which moves
with a stride size $s$ over the entire image. For example, for evaluating NeRF, people often use the kernel size $11 \times 11$, the stride size $1$, and the circular symmetric Gaussian weighting function $\bm{w} = \{w_i| i=1, 2, \dots, n\}$, with standard deviation of 1.5 samples, normalized to unit sum ($\sum_{i}w_i = 1$). The local statistics are then written as 
\begin{align}
\label{eq:localstat}
\mu_{a} = & \sum_{i=1}^{n} w_{i} a_{i}, \\
\sigma_{a} = & \left( \sum_{i=1}^{n} w_{i} (a_i - \mu_a)^{2} \right)^{\frac{1}{2}}, \\
\sigma_{ab} = & \left( \sum_{i=1}^{n} w_{i} (a_i - \mu_a) (b_i - \mu_b) \right)^{\frac{1}{2}}.
\end{align}

At each step, the local statistics and SSIM index are calculated within the local window. The final SSIM metric for evaluating NeRF is actually the mean SSIM (MSSIM) which is computed by averaging the SSIM indexes over each step.

\begin{table}[t]
\caption{Quantitative results of neural rendering of scenes in T$\&$T Intermediate. Model: DVGO.}
\label{table:nerftnt}
\begin{center}
\begin{small}
\resizebox{0.45\textwidth}{!}{%
\begin{tabular}{c|c|ccc}
\toprule
Scene &  Training & PSNR($\uparrow$) & SSIM($\uparrow$) & LPIPS($\downarrow$) \\
\midrule 
\multirow{2}{*}{M60}   & Standard & 17.49 & 0.647 & 0.501  \\
  & Multiplex & \textbf{17.71} & \textbf{0.652} & \textbf{0.483} \\ 
\midrule 
\multirow{2}{*}{Playground}   & Standard &  22.74 & 0.669 & 0.467 \\
  & Multiplex & \textbf{22.75} & \textbf{0.681} & \textbf{0.444} \\ 
\midrule 
\multirow{2}{*}{Train}   & Standard & 17.19 & 0.566 & 0.533 \\
  & Multiplex & \textbf{18.24} & \textbf{0.581} & \textbf{0.491}  \\
  \midrule 
\multirow{2}{*}{Truck}   & Standard & 22.01 & 0.704 & 0.386 \\
  & Multiplex & \textbf{22.44} & \textbf{0.730} & \textbf{0.338} \\ 
\bottomrule
\end{tabular}
}
\end{small}
\end{center}
\end{table}

\begin{table}[t]
\caption{Quantitative results of NeRF methods on T$\&$T. Model: DVGO.}
\label{table:dvgotntad}
\begin{center}
\begin{small}
\resizebox{0.45\textwidth}{!}{%
\begin{tabular}{c|c|ccc}
\toprule
Scene & Training &  PSNR($\uparrow$) & SSIM($\uparrow$) & LPIPS($\downarrow$)\\
\midrule 
\multirow{2}{*}{Scene 1}  & Standard & 21.80 & 0.759 & 0.243 \\
 & Multiplex & \textbf{22.16} & \textbf{0.783} & \textbf{0.191} \\  
 \midrule 
\multirow{2}{*}{Scene 2}  & Standard & 23.96 & 0.847 & 0.250  \\
 & Multiplex & \textbf{25.37} & \textbf{0.872} & \textbf{0.171} \\  
 \midrule 
\multirow{2}{*}{Scene 3}  & Standard & 18.64 & 0.697 & 0.272 \\
 & Multiplex  & \textbf{19.71} & \textbf{0.757} & \textbf{0.192} \\  
 \midrule 
\multirow{2}{*}{Scene 4}  & Standard & 25.27 & 0.800 & 0.179 \\
 & Multiplex & \textbf{25.55} & \textbf{0.823} & \textbf{0.151} \\  
 \midrule 
\multirow{2}{*}{Mean}  & Standard & 22.42 & 0.776 & 0.236 \\
 & Multiplex & \textbf{23.20} & \textbf{0.809} & \textbf{0.176} \\  
\bottomrule
\end{tabular}
}
\end{small}
\end{center}
\end{table}

\begin{table}[t]
\caption{Quantitative results of NeRF methods on T$\&$T. Model: NeRF.}
\label{table:nerftntad}
\begin{center}
\begin{small}
\resizebox{0.45\textwidth}{!}{%
\begin{tabular}{c|c|ccc}
\toprule
Scene & Training &  PSNR($\uparrow$) & SSIM($\uparrow$) & LPIPS($\downarrow$)\\
\midrule 
\multirow{2}{*}{Scene 1}  & Standard & 20.11 & 0.647 & 0.351\\
 & Multiplex & \textbf{22.34} & \textbf{0.714} & \textbf{0.302} \\
 \midrule 
\multirow{2}{*}{Scene 2}  & Standard & 23.09 & 0.774 & 0.349\\
 & Multiplex & \textbf{25.20} & \textbf{0.848} & \textbf{0.247} \\  
 \midrule 
\multirow{2}{*}{Scene 3}  & Standard & 17.22 & 0.523 & 0.486\\
 & Multiplex & \textbf{18.13} & \textbf{0.571} & \textbf{0.467} \\ 
  \midrule 
\multirow{2}{*}{Scene 4}  & Standard & 23.65 & 0.692 & 0.269\\
 & Multiplex & \textbf{24.88} & \textbf{0.767} & \textbf{0.198} \\ 
 \midrule 
\multirow{2}{*}{Mean}  & Standard & 21.02 & 0.659 & 0.364 \\
 & Multiplex & \textbf{22.64} & \textbf{0.725} & \textbf{0.304} \\
\bottomrule
\end{tabular}
}
\end{small}
\end{center}
\end{table}

\begin{table}[t]
\caption{Quantitative results of NeRF methods on T$\&$T. Model: TensoRF.}
\label{table:tensorftntad}
\begin{center}
\begin{small}
\resizebox{0.45\textwidth}{!}{%
\begin{tabular}{c|c|ccc}
\toprule
Scene & Training &  PSNR($\uparrow$) & SSIM($\uparrow$) & LPIPS($\downarrow$)\\
\midrule 
\multirow{2}{*}{Scene 1}  & Standard & 17.33 & 0.643 & 0.413\\
 & Multiplex & \textbf{22.67} & \textbf{0.779} & \textbf{0.216} \\
 \midrule 
\multirow{2}{*}{Scene 2}  & Standard & 24.44 & 0.847 & 0.240\\
 & Multiplex & \textbf{25.14} & \textbf{0.859} & \textbf{0.214} \\  
 \midrule 
\multirow{2}{*}{Scene 3}  & Standard & 12.15 & 0.342 & 0.761\\
 & Multiplex & \textbf{18.64} & \textbf{0.699} & \textbf{0.286} \\ 
  \midrule 
\multirow{2}{*}{Scene 4}  & Standard & 24.84 & 0.766 & 0.207 \\
 & Multiplex & \textbf{24.93} & \textbf{0.770} & \textbf{0.205} \\ 
 \midrule 
\multirow{2}{*}{Mean}  & Standard & 19.69 & 0.650 & 0.365 \\
 & Multiplex & \textbf{22.85} & \textbf{0.777} & \textbf{0.230} \\  
\bottomrule
\end{tabular}
}
\end{small}
\end{center}
\end{table}

\begin{table}[t]
\caption{Quantitative results of few-shot learning. We only keep eight training images from the Truck Scene, T$\&$T Intermediate. Model: DVGO.}
\label{table:fewshotnerf}
\begin{center}
\begin{small}
\begin{tabular}{c|c|ccc}
\toprule
Scene & Training  & PSNR($\uparrow$) & SSIM($\uparrow$) & LPIPS($\downarrow$)\\
\midrule 
\multirow{2}{*}{Truck}  & Standard & 11.37 & 0.343 & 0.704 \\
  & Multiplex & \textbf{13.43} & \textbf{0.372} & \textbf{0.610} \\   
\bottomrule
\end{tabular}
\end{small}
\end{center}
\end{table}

\begin{table}[t]
\caption{Quantitative results of neural rendering from sparse training images. Size indicates the portion of training samples kept from the original training dataset.}
\label{table:sparsenerf}
\begin{center}
\begin{small}
\begin{tabular}{c|c|ccc}
\toprule
Size & Training & PSNR($\uparrow$) & SSIM($\uparrow$) & LPIPS($\downarrow$)  \\
\midrule 
\multirow{2}{*}{20\%}  & Standard & 14.67 & 0.527 & 0.597 \\
  & Multiplex & \textbf{18.99} & \textbf{0.618} & \textbf{0.405} \\  
\midrule 
\multirow{2}{*}{40\%}  & Standard & 18.99 & 0.639 & 0.425  \\
 & Multiplex & \textbf{21.68} & \textbf{0.704} & \textbf{0.356}  \\  
\midrule 
\multirow{2}{*}{60\%}  & Standard &  21.07 & 0.683 & 0.396  \\
  & Multiplex & \textbf{22.11} & \textbf{0.719} & \textbf{0.347} \\  
\midrule 
\multirow{2}{*}{80\%}  & Standard & 21.74 & 0.686 & 0.386  \\
  & Multiplex &  \textbf{22.38} & \textbf{0.722} & \textbf{0.351} \\  
\bottomrule
\end{tabular}
\end{small}
\end{center}
\end{table}



\begin{table}[t]
\caption{Quantitative results of neural rendering from corrupted training images.}
\label{table:corruptednerf}
\begin{center}
\begin{small}
\begin{tabular}{c|c|ccc}
\toprule
 Noise Scale & Training & PSNR($\uparrow$) & SSIM($\uparrow$) & LPIPS($\downarrow$)  \\
\midrule 
\multirow{2}{*}{0.2}  & Standard & 21.36 & 0.663 & 0.453 \\
  & Multiplex & \textbf{21.94} & \textbf{0.686} & \textbf{0.420} \\  
\midrule 
\multirow{2}{*}{0.4}  & Standard & 18.20 & 0.584 & 0.569  \\
  & Multiplex & \textbf{20.89} & \textbf{0.636} & \textbf{0.494}  \\  
\midrule 
\multirow{2}{*}{0.6}  & Standard &  16.16 & 0.542 & 0.684  \\
 & Multiplex & \textbf{18.06} & \textbf{0.571} & \textbf{0.599} \\  
\bottomrule
\end{tabular}
\end{small}
\end{center}
\end{table}

\section{Supplementary Experimental Results}
\label{sec:suppexp}

In this section, we present supplementary experimental results.

We first present the quantitative results of DVGO on 4 scenes of T$\&$T Intermediate in Table \ref{table:nerftnt}.


We present the quantitative results of DVGO, NeRF, and TensoRF on each T$\&$T scenes in Tables \ref{table:dvgotntad}, \ref{table:nerftntad}, and \ref{table:tensorftntad}, respectively.

\textbf{Few-shot learning} We evaluate \textit{multiplex training} on a few-shot learning task, where we only keep eight training images. The few-shot learning quantitative results in Table \ref{table:fewshotnerf} support the significant advantage of \textit{multiplex training} via S3IM.

\textbf{Sparse inputs} We present the quantitative results of neural rendering from sparse training images in Table \ref{table:sparsenerf}.

\textbf{Robustness to corrupted images} We present the quantitative results of neural rendering from Corrupted Truck in Table \ref{table:corruptednerf}.

\end{document}